%% file: tacl2018v2-template.tex
\documentclass[11pt,a4paper]{article}
\usepackage[dvipsnames]{xcolor}
\usepackage{soul}
\usepackage{times,latexsym}
\usepackage{url}
\usepackage[T1]{fontenc}

\usepackage[acceptedWithA]{tacl2018v2}

\usepackage{graphicx} %
\usepackage{float}
\usepackage{subfig} %
\usepackage{booktabs} %
\usepackage{multirow} %
\usepackage{comment} %
\usepackage{amssymb} %
\usepackage{multicol}
\usepackage{dsfont} %
\usepackage{makecell} %
\usepackage{enumerate} %
\usepackage{xcolor}

\definecolor{xgreen}{rgb}{0.33,0.51,0.27}
\definecolor{xred}{rgb}{0.68,0.36,0.29}

\usepackage{covington} %

\usepackage{xparse}   %

\usepackage[utf8]{inputenc}
\usepackage{vcell}
\usepackage{array}
\usepackage{rotating}
\usepackage{tikz-dependency}

\usepackage{fnpct} %

\NewDocumentCommand{\rot}{O{60} O{1em} m}{\makebox[#2][l]{\rotatebox{#1}{#3}}}%

\usepackage{xspace,mfirstuc,tabulary}

\newif\iftaclinstructions
\taclinstructionsfalse %
\iftaclinstructions
\renewcommand{\confidential}{}

\newcommand{\instr}
\fi

\iftaclpubformat %

\else

\fi

\newcommand{\Task}{TNE}
\newcommand{\task}{NP Enrichment}
\newcommand{\taskf}{Text-based NP Enrichment}
\newcommand{\dataset}{NP Enrichment}

\title{The Missing Links: Definition, Dataset and Models for \\Text-based NP Enrichment}
\title{Text-based NP Enrichment:\\
Recovering preposition mediated relations between noun phrases}
\title{The Missing Link: Text-based NP Enrichment}
\title{Text-based NP Enrichment}

\author{Yanai Elazar$\thanks{~~Equal contribution.}$ \,\,
 Victoria Basmov$\footnotemark[1]$ \,\,
 Yoav Goldberg\,\,
 Reut Tsarfaty \\
Computer Science Department, Bar Ilan University \\
Allen Institute for Artificial Intelligence \\
{\tt  \{yanaiela,vikasaeta,yoav.goldberg,reut.tsarfaty\}@gmail.com}}

\date{}

\begin{document}
\maketitle
\begin{abstract}
Understanding the relations between entities denoted by NPs in a text is a critical part of human-like natural language understanding.
However, only a fraction of such relations is covered by standard NLP tasks and benchmarks nowadays.
In this work, we propose a novel task termed  
{\em text-based NP enrichment} (TNE), in which we aim to enrich each NP in a text with all the preposition-mediated relations --- either explicit or implicit --- that hold between it and other NPs in the text.
The relations are represented as triplets, each denoted by two NPs related via a preposition. 
Humans recover such relations seamlessly, while current state-of-the-art models struggle with them due to the implicit nature of the problem. We build the first large-scale dataset for the problem, provide the formal framing and scope of annotation, analyze the data, and report the results of fine-tuned  language models on the task, demonstrating the challenge it poses to current technology.
A webpage with a data-exploration UI, a demo, and links to the code, models, and leaderboard, to foster further research into this challenging problem can be found at: \url{yanaiela.github.io/TNE/}.
\end{abstract}

\section{Introduction}

\begin{figure}[t!]
    \centering
    \includegraphics[width=1.\columnwidth]{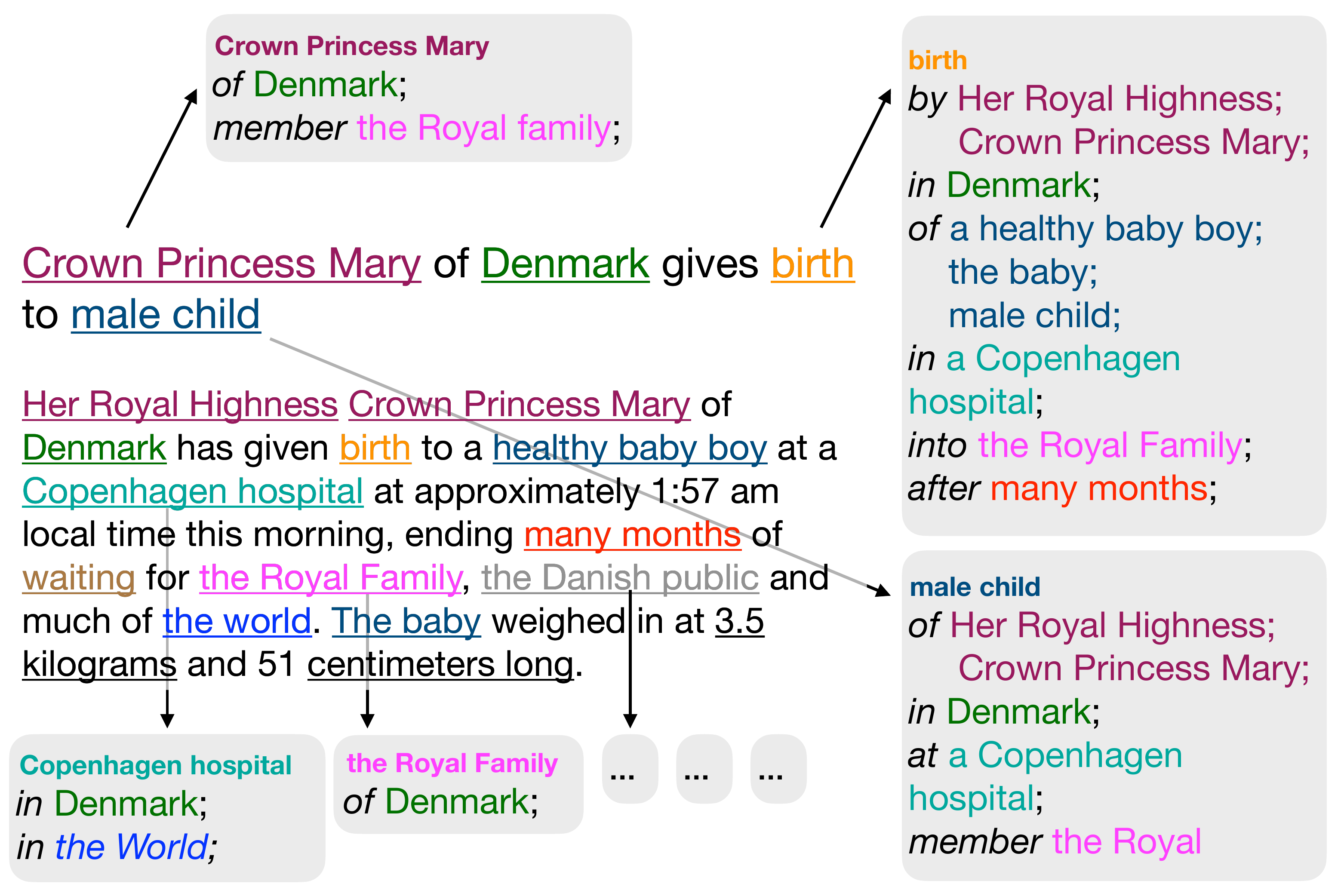}
    \caption{Preposition-mediated relations between NPs in a text. NPs with the same color designate the same entity (co-refer). 
    Gray boxes show all the preposition-mediated relations for a single NP anchor (some are indicated with ``...'' for brevity).
    This figure shows a title and a single short paragraph. The texts in our dataset span 3 paragraphs.}
    \label{fig:main}
    \vspace{-12mm}
\end{figure}

\begin{figure*}[t]
\centering
\includegraphics[width=1.\textwidth]{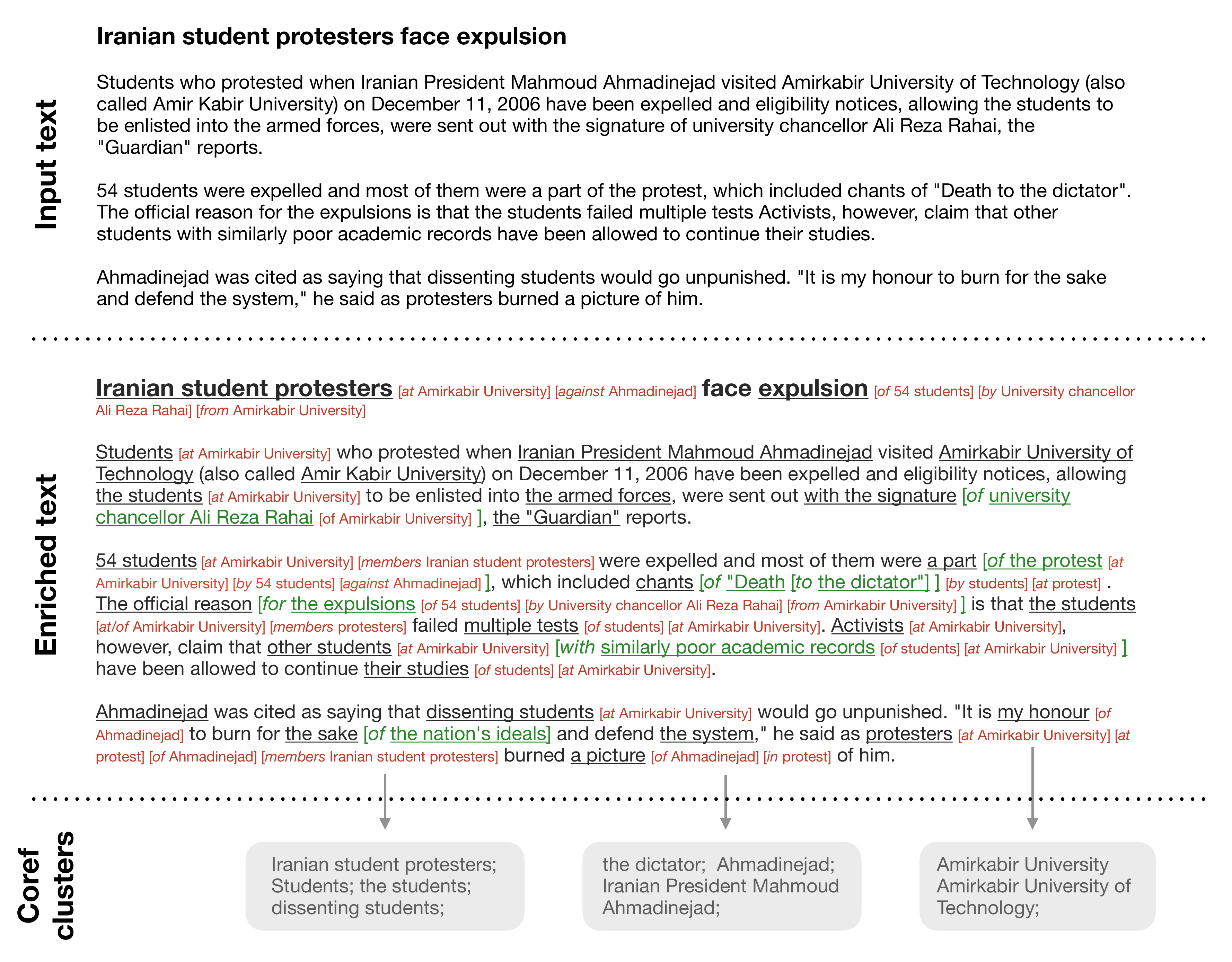}
\caption{NP-enriched document from the dataset. The title appears in a larger font, the NPs in the document are marked with \underline{underline}. The \textcolor{xgreen}{\underline{green}} NP-enrichments appear explicitly in the original text, while the \textcolor{xred}{red} do not, and are typically harder to infer. For brevity, each link in the text mentions only one of the NPs in a coreference cluster. The dataset has additional links to the other NPs in the cluster.}
\label{fig:complete}

\end{figure*}

A critical part of understanding a text is detecting the entities in the text, denoted by NPs, 
and determining the different semantic relations that hold between them.
Some semantic relations between NPs are explicitly mediated via verbs, as in (1):
	\noindent\begin{dependency}
    	\begin{deptext}
    	(1) \& \textit{Water} \& enters \& \textit{the plant} \& through the roots.\\
    	\end{deptext}
    	\depedge[edge height=0.5em]{2}{4}{enters}
	\end{dependency}
Much work in NLP addresses the recovery of such  verb-mediated relations (SRL) \cite{gildea2002automatic,srlbook}, either using pre-specified role ontologies such as PropBank or FrameNet \cite{palmer2005proposition,ruppenhofer2006framenet}, or, more recently, using natural-language-based representations (QA-SRL) \cite{he2015question,fitzgerald2018large}.
Another well-studied kind of semantic relations between NPs is that of \textit{coreference} \cite{muc1,conll2012}, where two (or more) NPs refer to the same entity.

Such NP-NP relations, that are either mediated by verbs (as in SRL or Relation Extraction) or form coreference relations, represent only a subset of the NP-NP relations that are naturally expressed in texts. %
Consider, for instance, the following  sentences:
	\noindent\begin{dependency}
    	\begin{deptext}
    	(2) \& \textit{A person} \& with \& \textit{brown eyes} \& crossed the street. \\
    	\end{deptext}
    	\depedge[edge height=0.5em]{2}{4}{with}
	\end{dependency}

	\noindent\begin{dependency}
    	\begin{deptext}
    	(3) Water enters \& \textit{the plant} \& through \& \emph{the roots}.\\
    	\end{deptext}
    	\depedge[edge height=0.5em]{4}{2}{of}
	\end{dependency}
	\noindent\begin{dependency}
    	\begin{deptext}
    	(4) \& I entered \& \textit{the room}, \&  \& \emph{the window} was  \& open.\\
    	\end{deptext}
    	\depedge[edge height=0.5em]{5}{3}{in}
	\end{dependency}
All of the above cases contain examples of NP-NP relations, where the type of relation can be expressed via an English preposition.
The preposition may be explicit in the text,
as in (2), where the relation \textbf{A person} \underline{with} \textit{blue eyes} is explicitly expressed, or they may be \emph{implicit} and left to the reader to infer, as in (3)-(4); in (3) readers easily  {infer} that \textbf{the roots} are \underline{of} \textit{the plant}. Likewise, in (4)  readers infer that  \textbf{the window} is \underline{in} \textit{the room}.\footnote{Here, both ``in'' and ``of'' are possible prepositions, but ``in'' is slightly more specific.}
Properly understanding the text means knowing that these relations hold, even when they are not explicitly stated in the utterance. Figure \ref{fig:main} shows additional examples.

These relations, both explicit and implicit, are indispensable for understanding the text. While human-readers infer these relations intuitively and spontaneously while reading, machine-readers generally ignore them. 
In this work, we thus propose a new NLU task in which we aim to recover all the preposition-mediated relations --- whether explicit or implicit --- between NPs that exist in a text. We call this task  \emph{Text-based NP Enrichment} or {\em \Task{}} for short.

The short examples~(2)--(4) that illustrate the phenomenon may not look  challenging to infer using current NLP technology. However, when we go beyond sentence level to document level, 
things become substantially more complicated. 
As we demonstrate in~\S\ref{sec:dataset-stats}, a typical 3-paragraph text in our dataset has an average of 35.8 NPs, which participate in an average of 186.7 preposition-mediated relations, the majority of which are implicit.
Figure \ref{fig:complete} shows a complete annotated document from our dataset.

The type of information recovered by the \task{} task complements well-established core NLP tasks such as entity typing, entity linking, coreference resolution, and semantic-role labeling \cite{Jurafsky+Martin:2009a}. We believe it serves as an important and much-needed building block for downstream applications that require text understanding, including information retrieval, relation extraction and event extraction, question answering, and so on. In particular, the \task{} task \emph{neatly encapsulates a lot of the long-range information} that is often required by such applications. 
Take for example a system that attempts to extract reports on police shooting incidents \cite{keith-etal-2017-identifying}, with the following challenging, but not uncommon, passage:\footnote{We thank Katherine Keith for this example.}
\begin{quote} \emph{Police officers spotted the butt of a handgun in Alton Sterling's front pocket and saw him reach for the weapon before opening fire, according to a Baton Rouge Police Department search warrant filed Monday that offers the first police account of the events leading up to \textbf{his fatal shooting}.} \end{quote}
Considering this shooting-event passage, %
an ideal coreference model will resolve \emph{his} to \emph{Alton Sterling's}, making the entity being shot local to the shooting event.  On top of that, an ideal \task{} model as we propose here will also recover:
\begin{quote}
\emph{fatal shooting [of Alton Sterling] [by Police officers [of Baton Rouge Police Department]]}
\end{quote}

\noindent making the shooter identity local to the shooting event as well, ready for use by a downstream event-argument extractor or machine reader. %

Of course, one could hope that a dedicated, end-to-end-trained shooting-events extraction model will learn to recover such information on its own. However it will require to pre-define the frame of {\em shooting} events, and it will require a substantial amount of training data to get it right (which often does not happen in practice). Focusing on {\em \taskf{}} %
provides an opportunity to learn a core NLU skill that does not focus on a pre-defined set of relations, and is not specific to a particular benchmark.
Finally, eyond its potential usefulness for downstream NLP applications, the \taskf{} task serves as a \emph{challenging benchmark for reading comprehension}, as we further elaborate in \S{\ref{sec:tne-understanding}}.

In what follows we formally define the {\em \taskf{}} task (\S\ref{sec:links-definition}) and its relation to {\em reading comprehension}  (\S\ref{sec:tne-understanding}), we describe a large-scale high-quality English TNE dataset we collected (\S\ref{sec:dataset}) and its curation procedure (\S\ref{sec:annotation}). We analyze the dataset (\S\ref{sec:dataset-stats}) and experiment with pre-trained language model baselines (\S\ref{sec:modeling}), achieving moderate but far-from-perfect success on this dataset (\S\ref{sec:neural}). We also conduct an analysis of the best model, showcasing the strenghts, weaknesses and open challenges of the best model (\S \ref{sec:res-analysis}). 
We then discuss the relation of TNE to other linguistic concepts, such as \emph{bridging} \cite{bridging}, \emph{relational nouns} \cite{partee1983a,relational-loebner,barker1995a} and \emph{implicit arguments} \cite{ruppenhofer-implicit-args,meye:anno04,gerber_implicit,cheng2019implicit} (\S\ref{sec:related}).
We finally conclude that, in contrast to those linguistic tasks, the \emph{\taskf{}} task is more exhaustive, sharply scoped, easier to communicate, and substantially easier to consistently annotate and use by non-experts.

\section{Text-based NP Enrichment (TNE)}
\label{sec:links-definition}

\paragraph{Task Definition} The Text-based NP Enrichment task is deceptively simple: for each ordered pair ($n_1$, $n_2$) of non-pronominal base-NP\footnote{We follow the definition of Base-NPs as defined by \citet{ramshaw-marcus-1995}: initial portions of non-recursive noun-phrases, including pre-modifiers such as determiners, adjectives and noun-compounds, but not including post-modifiers such as prepositional phrases and clauses. These are also known in the NLP literature as ``NP Chunks''.} spans in an input text, determine if there exists a preposition-mediated relation between $n_1$ and $n_2$, and if there is one, determine the  preposition that best describes their relation.\footnote{During annotation, we noticed that annotators often tried to express set-membership using prepositions, which resulted in awkward and unclear annotations. To remedy this, we found it effective to add an explicit ``member-of'' relation as an allowed annotation option. This significantly reduced confusion, increased consistency, and improved quality of the annotation. While not officially part of the task, we do keep these annotations in the final dataset.} %
The output is a list of tuples of the form ($n_i$, \underline{prep}, $n_j$), where $n_i$ is called the \emph{anchor} and $n_j$ is called the \emph{complement} of the relation. Figure \ref{fig:complete} shows an example of text where each NP $n_1$ is annotated with its (\underline{prep}, $n_2$) NP-enrichments.

Despite the task's apparent simplicity, the underlying linguistic phenomena are quite complex, and range from simple syntactic relations to relations that require pragmatics, world-knowledge and common-sense reasoning. Performing well on the task suggests a human-like level of understanding. Notably, human readers detect most of the relations almost subconsciously when reading, while some of the relations require an extra conscious inference step.
\paragraph{Example}
Consider the following text: \\[0.5em] 
(5) \label{ex:running} \emph{Adam}'s \emph{father} went to meet \emph{the teacher} at \emph{his school}.\\[0.5em]
\noindent In this utterance, there are four non-pronominal base-NPs: ``Adam'', ``father'', ``the teacher'' and ``his school'', which makes 12 possible pairs: (\textbf{Adam}, \textit{father}), (\textbf{Adam}, \textit{the teacher}), $\dots$, (\textbf{his school}, \textit{the teacher}).

The preposition-mediated relations to be recovered in this example are:\footnote{We note that some of these are ambiguous. For example, it is not 100\% certain that the teacher is indeed ``the Teacher of Adam''. E.g., it could be a teacher of Adam's sister. Yet, without further information, many if not most readers will interpret it as the teacher of Adam. This kind of ambiguity is an inherent property of language, and---like in many other datasets, cf. NLI--- we deliberately opted for wide-coverage over preciseness: we are interested in what a ``typical human'' might infer as a relation, and not only 100\% certain ones.}
\begin{enumerate}[i]
    \item (\textbf{father}, \underline{of} \textit{Adam})
    \item (\textbf{the teacher}, \underline{of}, \textit{Adam})
    \item (\textbf{the teacher}, \underline{at}, \textit{his school})
    \item (\textbf{his school}, \underline{of}, \textit{Adam})
\end{enumerate}
 The first items are anchors, and the latter ones are the complements.

\paragraph{Order} The order of appearance of NPs within the text does not matter: for a given pair of NPs $n_1$ and $n_2$, we consider both ($n_1$, $n_2$) and ($n_2$, $n_1$) as potential relation candidates, and it is possible that both relations will hold (likely with different prepositions). The only restriction is that an NP span cannot relate to itself. For a text with $k$ NPs, this results in $k^2-k$ candidate pairs.

\paragraph{Scope} 
In terms of the annotated relations, we are interested in the set of semantic relations that can be expressed 
in natural language 
via the use of a preposition. 
This identifies a rich, cohesive and well-scoped set of NP-NP relations that are not mediated by a verb and are not coreference relations.
Importantly, we restrict ourselves to NPs that are mentioned in the text, excluding relations with NPs that reside in some text-external shared context. For example, consider the sentence: ``\emph{The president discussed the demonstrations near the border}''. Here, the NPs ``{\em the president}'', ``{\em the border}'' and ``{\em the demonstrations}'' are all under-determined, and, to be complete, should relate to other NPs using preposition-mediated relations: \textbf{president} [\underline{of} \textit{Country}]; \textbf{border} [\underline{of} \textit{CountryX}] [\underline{with} \textit{CountryY}]; \textbf{demonstration} [\underline{by} \textit{some-group}] [\underline{about} \textit{some-topic}]. However, as these complement NPs do not appear in the text, we do not consider them to be part of the TNE task.

\paragraph{The Use of Prepositions as Semantic Labels} 
While the relations we identify between NPs can be expressed using prepositions, one could argue that using prepositions as semantic labels is not ideal, due to their inherent ambiguity \cite{pssts,psstcorpus,Schneider2018ComprehensiveSD,Gessler2021SupersenseAS}: indeed a preposition such as \emph{for} has multiple senses, and can indicate a large set of semantic relations ranging from \textsc{Beneficiary} to \textsc{Duration}. 

We chose to use prepositions as relation labels, despite this ambiguity.  
This follows a line of annotation work that aims to express semantic relations using natural language \cite{fitzgerald2018large,roit-etal-2020-controlled, klein-etal-2020-qanom, pyatkin2020qadiscourse}, as opposed to works that used formal linguistic terms, traditionally relying on expert-defined taxonomies of semantic roles and discourse relations. The aforementioned works   label  predicate-argument relations using restricted questions. In the same vein, we label nominal relations using prepositions. %

We argue that the preposition-based labels are useful for humans and machines alike: humans can easily understand the task (both as annotators and --- perhaps more importantly --- as consumers), and current machine learning models are quite effective with implicitly dealing with prepositions ambiguity.\footnote{For example, in the SQUAD question-answering dataset \cite{squad}, 15\% of all questions either begin or end with a preposition, and 30\% of all answer spans directly follow a preposition, requiring the models to deal with their ambiguous nature in order to perform well. The high accuracy scores obtained on the SQUAD dataset indicates that the models indeed succeed in the face of ambiguity. Relation-extraction tasks also work to a large extent around prepositional phrases, and manage to effectively extract relations.} Moreover, while the prepositions themselves are ambiguous, the (\textbf{NP}, \underline{prep}, \textit{NP}) triplet provides context which is, in many cases, sufficient to disambiguate the coarse-grained preposition sense.

We find that the preposition-based annotation has the following advantages: it clearly scopes the task with respect to the kinds of relations that are contained in it; and it is expressive, capturing a large class of interesting semantic relations. On top of that, the task and the corresponding relation-set is easy to explain to both human annotators (thus allowing to obtain high levels of agreement) and to human consumers of the model (allowing wider adoption, as the task and its output does not require special training to understand). Finally, the output can be easily fed into existing NLP systems, which already deal to a large extent with the inherent ambiguities of prepositions and prepositional phrases.

To conclude, we argue that despite the  ambiguities of prepositions, they allow us to  obtain a meaningful set of {\em typed} semantic links between NPs, which are well understood by people and can be effectively processed by NLP models. While the annotation can be refined to include a fine-grained sense annotation for each link, e.g., via a scheme as that of \citet{Schneider2018ComprehensiveSD}, we leave such an extension to future work.

\paragraph{Coreference Clusters}
A common relation between NPs is that of \emph{identity}, a.k.a.\ a {\em coreference} relation, where two or more NPs refer to the same entity. How do coreference relations relate to the \task{} task? While the \task{} task so far is posed as inferring prepositional relations between NPs, in actuality the prepositional relations hold between an NP and a coreference cluster. Indeed, if there is a prepositional relation prep($n_1$, $n_2$), and a coreference relation coref-to($n_2$, $n_3$), we can immediately infer the link prep($n_1$, $n_3$).\footnote{Note that the converse does not hold: prep($n_1$, $n_2$), coref-to($n_1$, $n_3$) does not necessarily entail prep($n_3$, $n_2$). Consider for example: ``\emph{The race began. John, the organizer,  pleased}''. While \textit{John} and \textit{the organizer} are coreferring, the relation \textbf{organizer} \underline{of} \textit{the race} holds, while \textbf{John} \underline{of} \textit{the race} does not.  %
This is because \textit{John} and \textit{the organizer} are two different senses for the same reference, and the relation holds only for one of the senses (cf.\ \citet{Frege60}). Putting it differently, when \textit{John} and \textit{organizer} serve as predicates, their selectional preferences are different despite them coreferring. Such examples are common, consider also 
\emph{``John is Jenny's father, Mary's husband''} where \textbf{father} \underline{of} \textit{Jenny} holds, while \textbf{husband} \underline{of} \textit{Jenny} doesn't. Similarly, \textbf{husband} \underline{of} \textit{Mary} holds, while \textbf{father} \underline{of} \textit{Mary} doesn't.}
We make use of this fact in our annotation procedure, and the dataset includes also the coreference information between all NPs in the text.
Indeed, for brevity, Figure \ref{fig:complete} shows only a subset of the relations, indicating for each anchor NP only a single complement NP from each coreference cluster. Some of the coreference clusters are shown at the bottom of the Figure. Note that the coreference clusters are not part of the task's input or expected output.

\paragraph{Formal Dataset Description}
An input text is composed of tokens $w_1, ..., w_t$, and an ordered set $N=n_1,...,n_k$ of base-NP mentions.
The underlying text is often arranged into paragraphs, and may also include a title. 
A base-NP mention, also known as NP chunk, is the smallest noun phrase unit that does not contain other NPs, prepositional phrases or relative clauses.\footnote{We use an automatic parser to obtain such base-NPs, using spaCy's parser \cite{spacy}.} It is defined as a contiguous span over the text, indicated by start-token and end-token positions (e.g., \texttt{(3, 5) ``the young boy''}).
The output is a set $R$ of relations of the form ($n_i$, \underline{prep}, $n_j$), where $i \neq j$ and \underline{prep} is a preposition (or a set-membership symbol). Each text is also associated with a set $C$ of non-overlapping coreference clusters, where each cluster $c \subseteq N$ is a non-empty list of NP mentions. The set of clusters is {\em not} provided as input, but for correct sets $R$ it holds that $\forall n_{j'} \in c(n_j)$, ($n_i$, \underline{prep}, $n_j$) $\in R \Rightarrow$ ($n_i$, \underline{prep}, $n_{j'}$) $\in R$,   where $c(n_j) \in C$ is the cluster containing $n_j$.

\paragraph{Completeness and Uniformity}
The kinds of preposition-mediated relations we cover originate from different linguistic or cognitive phenomena, and some of them can be resolved by employing different linguistic constructs.
For example, some within-sentence relations can be extracted deterministically from dependency trees, e.g., by following syntactic prepositional attachment.
Other relations can be inferred based on pronominal coreference (e.g., ``his school [of Adam]'' above can be resolved by first resolving ``his'' to ``Adam's'' via a coreference engine, and then normalizing ``Adam's school'' $\rightarrow$ ``school of Adam''). Many others are substantially more involved. We deliberately chose not to distinguish between the different cases, and expose all the relations to the user (and to the annotators) via the same uniform interface. This approach also contributes to the practical usefulness of the task: instead of running several different processes to recover different kinds of links, the end-user will have to run only one process to obtain them all.

\paragraph{Evaluation Metrics}

Our main metrics for evaluating NP enrichment tasks are precision, recall, and F1 on the recovered triplets (links) in the document. For analysis, we also report two additional metrics: precision/recall/F1 on unlabeled links (where the preposition identity does not matter), and accuracy of predicting the right preposition when a gold link is provided. We break this last metric into two quantities: accuracy of predicting the preposition for gold links that were recovered by the model, and accuracy of prepositions for gold links that were not recovered.

\section{TNE as a Reading Comprehension Benchmark}
\label{sec:tne-understanding}
While {\em reading comprehension} (RC) and {\em question answering} (QA) are often used interchangeably in the literature, measuring the reading comprehension capacity of models via question answering, as implemented in  benchmarks such as SQuAD \cite{squad}, BoolQ \cite{boolq} and  others, has several well-documented problems \cite{dunietz2020test}. We  argue that the TNE task we propose herein  has  properties that make it appealing  for assessing RC, more than QA is.

First, benchmarks for extractive (span-marking) QA are sensitive to the span-boundary selection, on the other hand, benchmarks for yes/no, multiple choice or generative questions can in principle be answered in a way which is completely divorced from the text.
On a more fundamental level, all QA benchmarks are very sensitive to lexical choices in the question and its similarity to the text. Furthermore, QA benchmarks rely on human authored questions that are easy to solve based on surface artifacts.
Finally, in many cases, the existence of the question itself provides a huge hint towards the answer \cite{kaushik2018much}.

The underlying cause for all of these issues is that QA-based setups do not measure the comprehension of a \emph{text}, but rather comprehending a (\textit{text}, \textit{question}) {pair}, where the question adds a significant amount of information, focuses the model on  specific aspects of the text, and exposes the evaluation to  biases and artifacts.
The reliance on the human-authored questions makes QA a bad format for measuring  ``text understanding'' 
--- we are likely measuring something else, such as the ability of the model to discern patterns in human question-writing behavior.

The TNE task we define side-steps all the above issues. %
It is based on the text alone, without revealing additional information not present in the text. The exhaustive nature of the task entails looking both at positive instances (where a relation exists) and negative ones (where it doesn't), making it harder for models to pick up shallow heuristics. %
We don't reveal information to a model, beyond the information that the two NPs appear in the same text.
Finally, the list of NPs to be considered is pre-specified,  isolating the problem of \emph{understanding the relations} between NPs in the text from the much easier yet intervening problem of \emph{identifying NPs} and \emph{agreeing on their exact spans}. %

Thus, we consider TNE a less biased and less gameable measure of RC than QA-based benchmarks. Of course, the information captured by TNE is limited and does not cover all levels of text understanding. %
Yet, performing the task correctly entails a non-trivial comprehension of texts, which human readers do as a byproduct of reading. 

\section{Text-based NP Enrichment Dataset}
\label{sec:dataset}

We collect a large-scale \Task{} dataset, consisting of 5.5K documents in English (3,988 train, 500 dev, 500 in-domain test, and 509 out-of-domain test). It covers about 200K NPs and over 1 million NP relations.
The main domain is WikiNews articles, and the out-of-domain (OOD) texts are split evenly between reviews from IMDB, fiction from project Gutenberg, and discussions from Reddit. 

Each annotated document consists of a title and  3 paragraphs of text, and contains a list of non-pronominal base-NPs (most identified by SpaCy \cite{spacy}\footnote{v.3.0.5, model \emph{en\_core\_web\_sm}.} 
but some added manually by the annotators), a list of coreference clusters over the NPs, and a list of NP-relations that hold in the text. Each relation is a triplet consisting of two NPs from the NP list, and a connecting element which is one of 23 prepositions (displayed in Table \ref{tab:prepositions})\footnote{The set was initiated with the 20 most common prepositions in English, and we added three additional prepositions that were requested during the initial annotation phase.} or a ``member(s) of'' relation designating set-membership.  The list of  NP relations  is {\em exhaustive}, and aims to cover all and only valid NP-NP relations in the document. %

\input{tables/prepositions}

\section{Data Annotation and Curation}
\label{sec:annotation}

\subsection{Annotation Procedure}

We propose a manual annotation procedure for collecting a large-scale dataset for the \Task{} task.
Considering all $k^2-k$ NP pairs (with an average $k$ of 35.8 in our dataset) is tedious, and, in our experience, results in mistakes and inconsistencies. In order to reduce the size of the space and improve annotation speed, quality, and consistency, we opted for a \textbf{two-stage process}, where the first stage includes the annotation of coreference clusters  over mentions, and the second stage involves \task{} annotation over the clusters from the first stage. We find that this two-stage process dramatically reduces the number of decisions that need to be taken, and also improves recall and consistency by reducing the cognitive load of the annotators,  focusing them on a specific mode at each stage. We hereby describe the different stages.

\begin{figure}[t!]
\centering

\subfloat[][Coreference clusters collection interface.]{\includegraphics[width=1.\columnwidth, trim={0 1.5cm 0 0},clip]{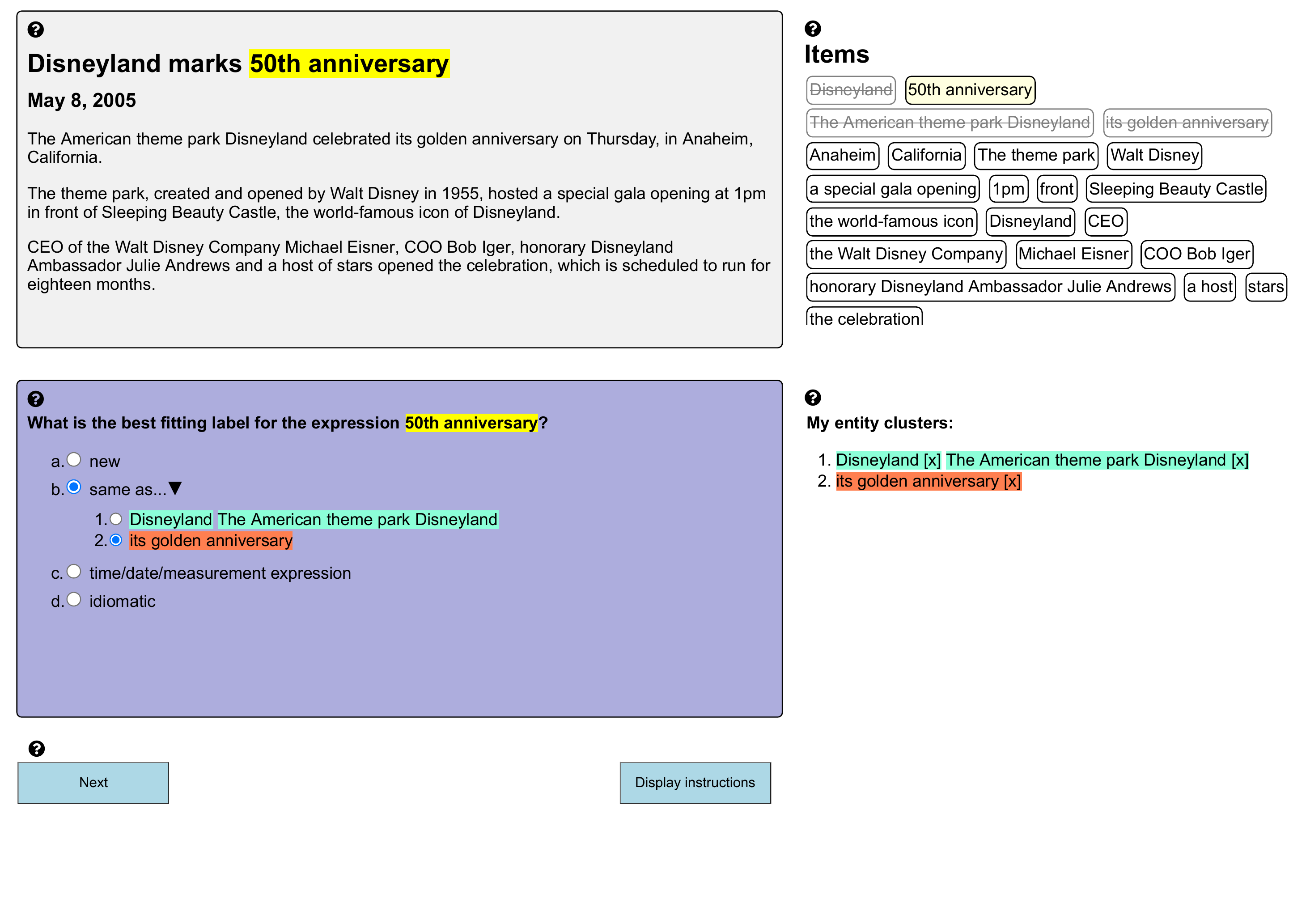}\label{fig:coref-annotation}} \\\hrule
\subfloat[][\task{} data collection interface.]{\includegraphics[width=1.\columnwidth, trim={0 1.5cm 0 0},clip]{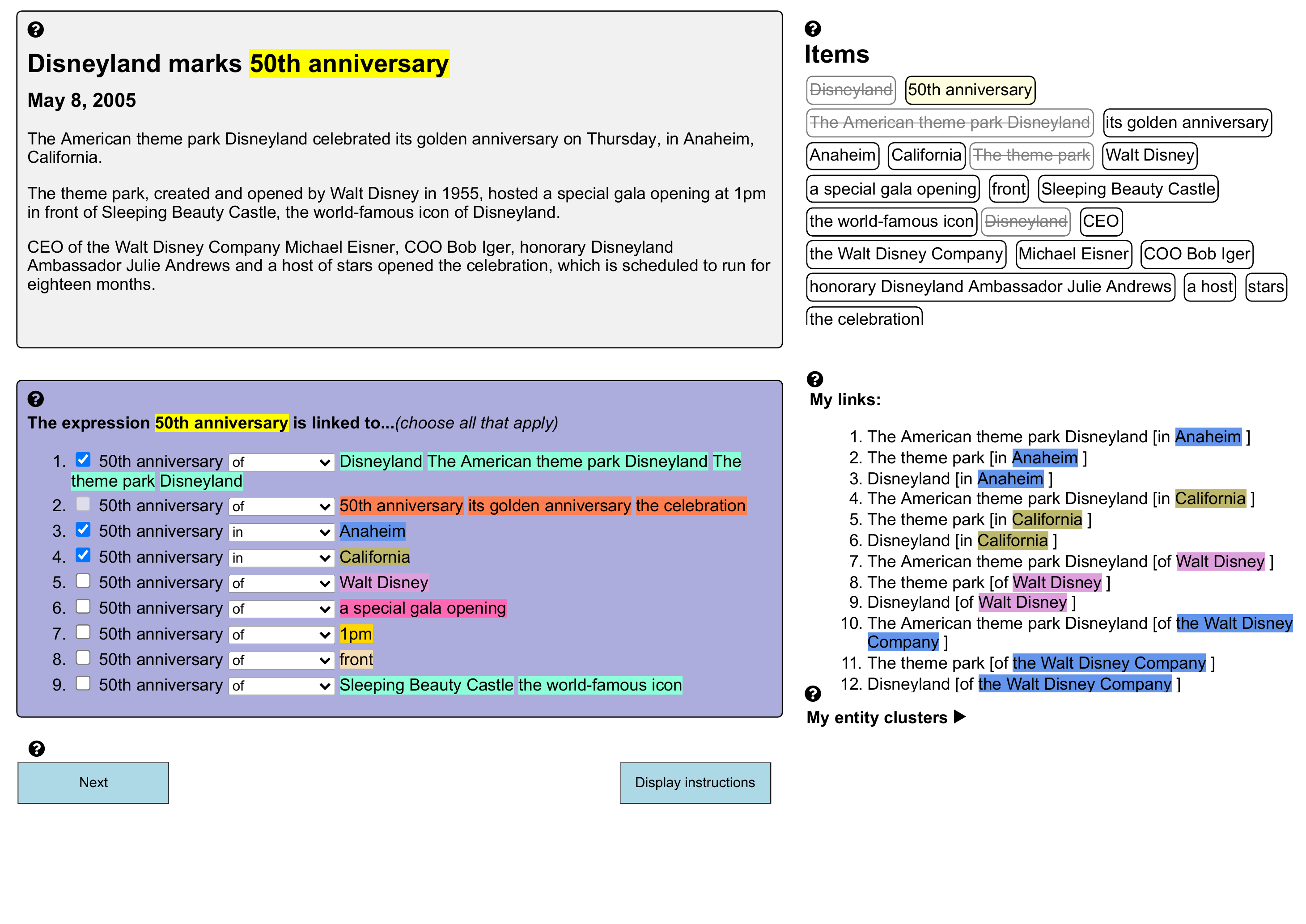}\label{fig:tne-annotation}} \\\hrule
\subfloat[][\task{} consolidation interface.]{\includegraphics[width=1.\columnwidth, trim={0 1.5cm 0 0},clip]{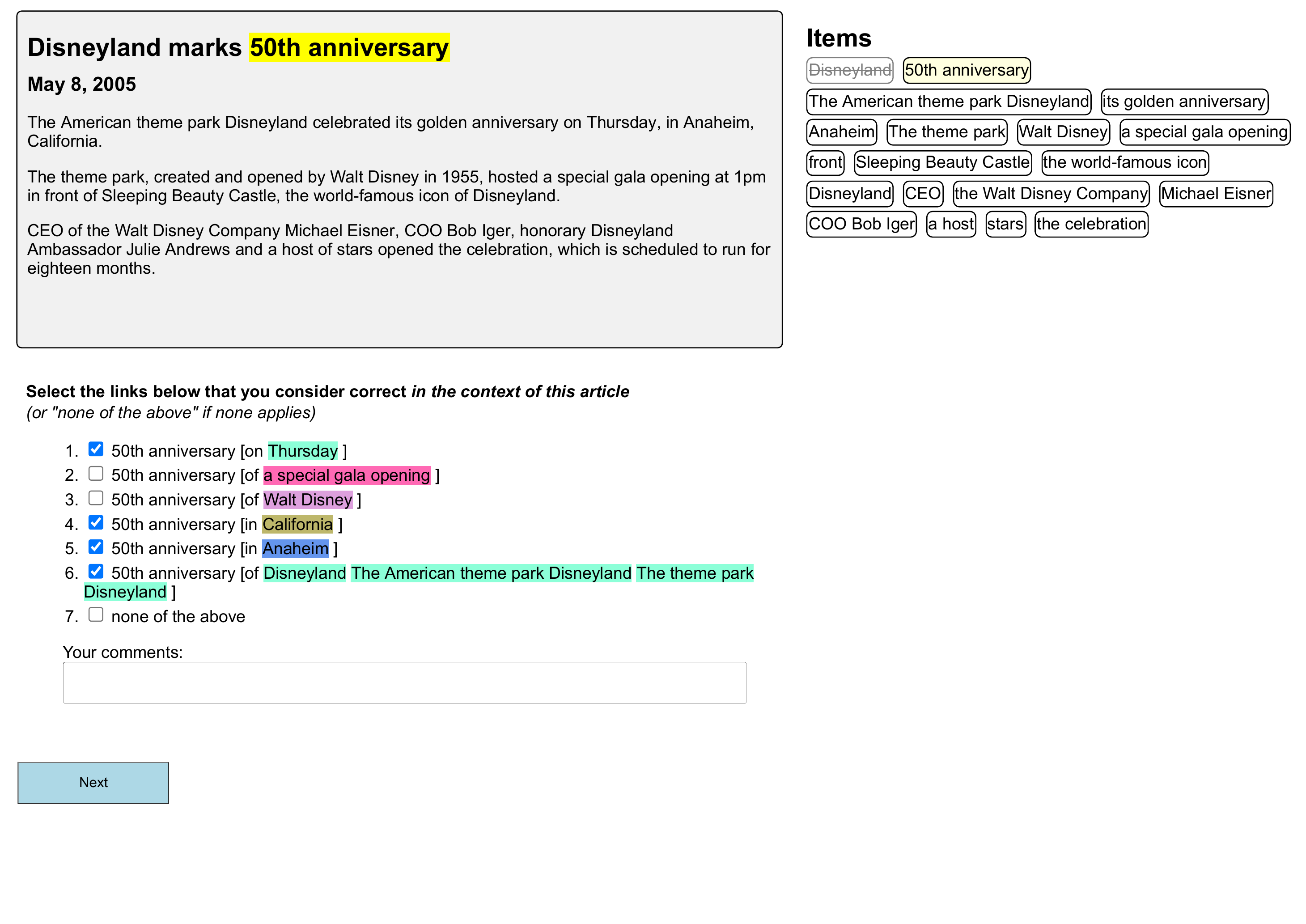}\label{fig:consolidation-annotation}} \\\hrule

\caption{Interfaces of the annotation steps.}
\label{fig:annotation_interface}
\vspace{-16mm}
\end{figure}

\paragraph{Stage 1: Annotating Coreference Clusters}

We start by collecting coreference clusters, as well as discarding non-referring NPs, that are ``irrelevant'' for the next stage (such as time-expressions). 
We created a dedicated user-interface to facilitate this procedure (Figure \ref{fig:coref-annotation}).
The annotators go over the NPs in the text in order, and, for each NP, indicate if it is (a) a new mention (forming a new cluster); (b) ``same as'' (coreferring to an entity in an existing cluster initiated earlier); (c) a time or measurement expression; (d) an idiomatic expression. At each point, the annotators can click on a previous NP to return to it and revise their decisions.

The OOD and documents from the test-set were annotated by two annotators for measuring agreement. They were then consolidated by one of the paper's authors for high quality annotations.

\paragraph{Stage 2: Annotating NP-relations}

The second step is the \task{} relation annotation. The annotators are exposed to a similar interface (Figure \ref{fig:tne-annotation}). For each NP, they are presented with all the coreference clusters, and must indicate for each cluster if there is a preposition-mediated relation between the NP and the cluster.

For this stage, all documents are annotated by two annotators and undergo a consolidation step.
The \textbf{consolidation} over the two annotators is performed by a third annotator, that did not see the document before. This annotator is presented with the interface shown in Figure \ref{fig:consolidation-annotation}. The consolidator sees all the relations created by the two preceding annotators, and decides which of them are correct.\footnote{We measure the agreement of this step by an additional annotators that consolidate 10\% of the documents, and report the agreement between the two consolidators.}\footnote{In this stage, two links with identical NPs can be chosen with different prepositions (e.g. Ex. (4)). This may increase the number of possible relations in a given document from $k^2 - k$ possible pairs to $(k^2 - k) * p$, where $p$ is the number of considered prepositions. However in practice, having more than two prepositions for the same NP pairs is not common, and two prepositions occur in 11.6\% of the test-set. For simplicity, in this work, we consider a single preposition for each NP pair, but the collected data may contain two prepositions for some pairs.}

\subsection{Annotators}
We trained and qualified 23 workers on the Amazon Mechanical Turk (AMT) platform, to participate in the coreference, NP relations, and consolidation tasks. We follow the controlled crowdsourcing protocol suggested by \citet{roit-etal-2020-controlled,pyatkin2020qadiscourse} giving detailed instructions, training the workers, and providing them with ongoing personalized feedback for each task. 

We paid 1.5\$, 2.5\$, and 1.5\$ for each HIT in the coreference, NP-relations, and consolidation tasks respectively. The price for the NP-relations task was raised to 2.7\$ for the test and out-of-domain subsets.
We additionally paid bonus payments on multiple occasions. Overall, we aimed at paying at least the minimum wage in the U.S.

\subsection{Inter-annotator Agreement}
\label{sec:quality}
We report the agreement scores for the coreference and the consolidated relation annotations. The full results, broken by split are reported in Table \ref{tab:agreement}.
The IPrep-Acc and UPrep-Acc metrics measure the preposition-only agreement (whether the annotators chose the same preposition for a given identified NP-pair), and are discussed in \S\ref{sec:quantitative-analysis}.

\input{tables/agreement}

\paragraph{Coreference}
We follow \citet{cattan-coref-eval} and evaluate the coreference agreement scores after filtering singleton clusters. We report the standard CoNLL-2012 score \cite{conll2012} that combines three coreference metric scores. The in-domain test score\footnote{To reduce costs and time, we did not collect double annotation for the train split, thus we cannot report agreement on it.} is 82.1, while in the OOD the score is 77.1. 
For comparison with the most dominant coreference dataset, OntoNotes \cite{ontonotes}, which only reported the MUC agreement score \cite{muc}, we also measure the MUC score on our dataset. The MUC score on our dataset is 83.6, compared to 78.4-89.4 in OntoNotes, depending on the domain \cite{conll2012}.
It is worth noting that on the Newswire domain of OntoNotes \cite{ontonotes} (the domain that is most similar to ours) the score is 80.9,  which indicates a high quality of annotation in our corpus.
We expect the quality of our final coreference data to be even higher due to the consolidation step that was done by an expert on the test set and OOD splits.%

\paragraph{NP-relations}
Next, we report agreement scores on the NP-relations consolidation annotation, which were measured on 10\% of all the annotations. We use the same metrics for the \task{} task (\S \ref{sec:links-definition}) and use one of the annotations as gold, and the other as the prediction. Thus we only report accuracy and F1 scores (the precision and recall are symmetric depending on the role of each document).
The Relation-F1 scores for the train and test are 89.8 and 94.4 respectively, while for the OOD it is 88.9. %
The preposition scores are almost perfect in all splits, with an average of 99.9 when the annotators agree on the link and 100.0 when they don't.
Finally, the F1 scores also differ between splits: 89.6, 94.4, and 88.6 for the train, test, and OOD, respectively, but are overall high.

\section{Dataset Statistics and Analysis}
\label{sec:dataset-stats}

\input{tables/stats}

We report  statistics of the resulting \dataset{} dataset, and summarize them in Table \ref{tab:analysis}.
Overall, we collected 5,497 documents, with per-document averages of 35.8 NPs, 5.2 non-singleton coreference clusters, and 186.7 NP-relations. The average number of tokens in a document is 163.3 tokens, where the largest document has 304 tokens.

\begin{figure}[t!]
\centering

\includegraphics[width=1.\columnwidth]{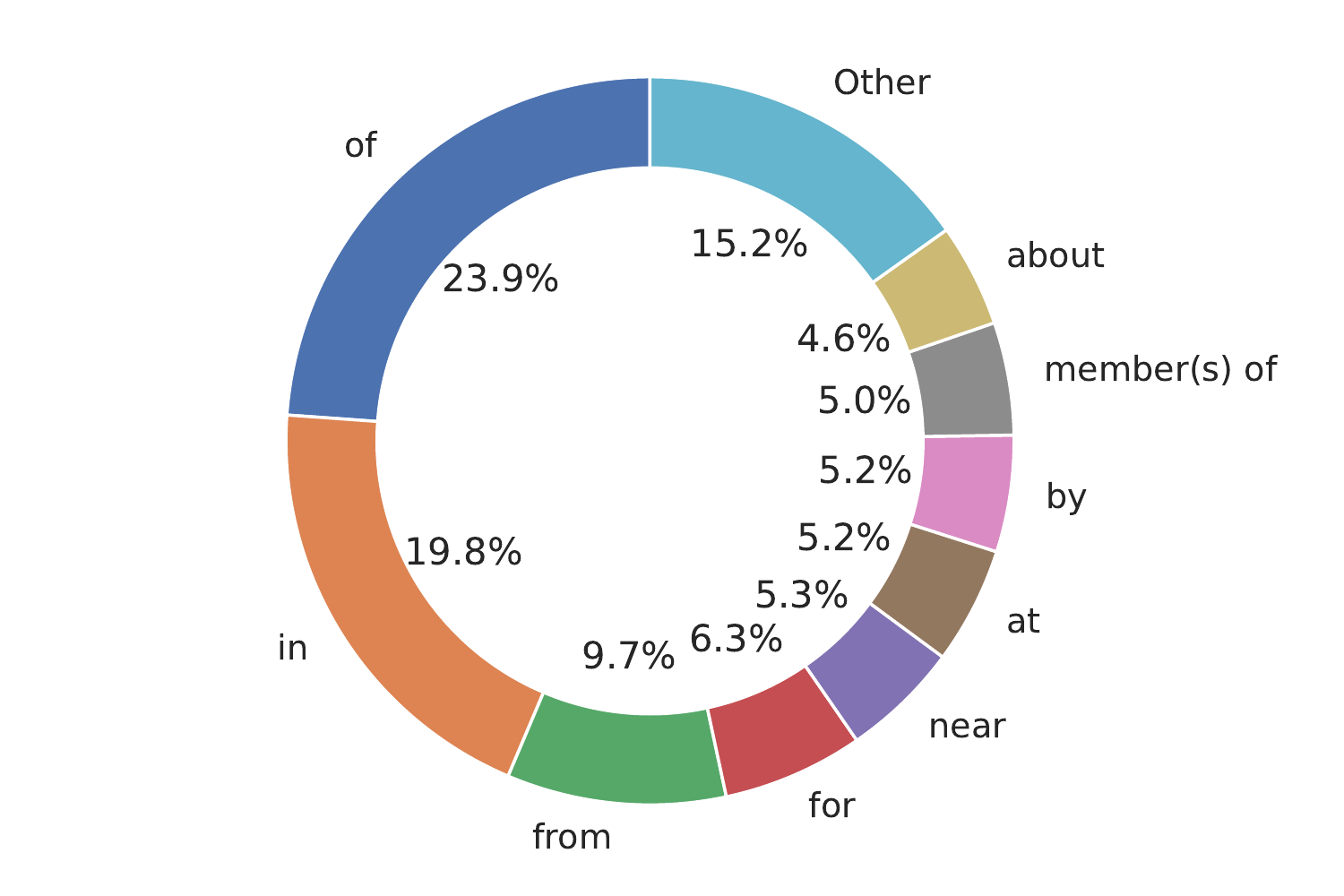}

\caption{Distribution of the prepositions in the \dataset{} test set.}
\label{fig:prepositions}
\vspace{-10mm}
\end{figure}

\paragraph{Distribution of Prepositions}
We analyze the prepositions in the relations we collected. We aggregate the prepositions of the test set from all relations and present their distribution in Figure \ref{fig:prepositions}. We only show prepositions that appear at least in 4\% of the data, and the rest are aggregated together into the \textit{Other} label.
The most common preposition is \textit{of}, followed by \textit{in}, which constitute 23.9\% and 19.8\% of the prepositions in our data respectively. The rest of the prepositions are used much less frequently, with \textit{from} and \textit{for} appearing in 9.7\% and 6.3\% of the prepositions respectively. The least used preposition is \textit{into}, which appears in 0.07\% of the prepositions.

\paragraph{NP-relations}

We provide some statistics that shed light on the nature of the preposition-mediated NP-NP relations in the annotated data.

First, we measure the \emph{surface distance} between NPs in the relations, in terms of token counts between the \textit{anchor} and the \textit{complement}. We found the average distance to be 53.7 tokens, indicating an average large distance between two NPs, which demonstrates the task's difficulty.
\textit{Backward-relations} (as opposed to \textit{forward-relations}) are relations where the \textit{complement} appears before the \textit{anchor}. 56.7\% of the relations are backward.
Sometimes, the string ``\textbf{anchor} \underline{preposition} \textit{complement}'' appears directly in the text. We call these cases \textit{Surface-Form}.  For instance in Ex (5) (\textbf{the teacher} \underline{at} \textit{his school}) is a \textit{Surface-Form} relation. We computed the percentage of such relations in the data and found only 3.9\% of them to be of such type.
We also relax this definition and search for the preposition following the complement in a window size of 10 from the anchor, which we call \textit{Surface-Form+}. The percentage of such cases remains low: 6.0\% of the links.
\textit{Symmetric} relations are two relations between the same two NPs, that differ in direction (and potentially the preposition). For instance in Figure \ref{fig:error-analysis-text}, the following links are symmetric (\textbf{website}, \underline{of}, \textit{the owners}) and (\textbf{the owners}, \underline{of}, \textit{website}). On average, there are 10.9 such symmetric relations in a document.
Finally, \textit{transitive} relations are sets of three NPs, \textit{a}, \textit{b} and \textit{c} that include relations between  $(a, b)$, $(b, c)$ and $(a, c)$ (the preposition identity is not relevant). %
We found an average of 97.3 transitive relations per document in total.

\paragraph{Explicit vs.\ Implicit NP Relations} 
Next, we analyze the composition of the relations in the dataset, as to whether these relations are implicit or explicit. 
While there is no accepted definition of explicit-implicit distinction in the literature \cite{Carston,Jarrah2016ExplicitimplicitDA}, here we adapt a definition originally used by \citet{cheng2019implicit} for another phenomenon, implicit arguments:\footnote{Implicit arguments ``are not syntactically connected to the predicate and may not even be in the same sentence'' \cite{cheng2019implicit}.} in an \textit{implicit} relation the \textit{anchor} and the \textit{complement} are not syntactically connected to each other and might not even appear in the same sentence. This implies, e.g., that any inter-sentential relations are implicit\footnote{84.6\% of all the links in our dataset are inter-sentential.}, while relations within one sentence can be either implicit or explicit. 
We sample three documents from the test-set, containing 590 links in total, and count the number of relations of each type.
Our manual analysis reveals that 89.8\% of the relations are implicit.

\paragraph{Bridging vs.\ TNE} \textit{Bridging} has been  extensively studied in the past decades, as we discuss in \S\ref{sec:related}. Here, we explore how many of the relations we collected correspond to the definition of \textit{bridging}.
We use the same three documents from the analysis described above, and follow the annotation scheme from ISNotes1.0 \cite{isnotes}\footnote{\url{https://github.com/nlpAThits/ISNotes1.0/blob/master/doc/release_annotation_scheme.pdf}} to annotate them for \textit{bridging}. We found that 15 out of the 590 links (2.5\%) in these  documents are \textit{bridging} links (i.e., meet the criteria for bridging defined in  ISNotes). These three documents contain 104 NPs, i.e., the ratio of bridging links per NP is 0.14. While the ratio is small, it is larger than the ratio in ISNotes which contains 663 bridging links out of 11K annotated NPs \cite{hou-etal-2013-global}, i.e., 0.06 bridging links per NP.

\section{Deterministic Baselines}
\label{sec:modeling}

We explore multiple deterministic baselines, that should expose regularities in the data that models may use (and therefore may result in an easy to solve dataset), and provide further insights about our data.
In these baselines we focus on detecting valid anchor/complement pairs, without considering the preposition's identity.

\paragraph{Title Link}
This baseline considers one of the title's NPs as the \textit{complement} for each NP in the text. We experiment with three variants: \textit{Title-First}, \textit{Title-Last} and \textit{Title-Random} which use the first, last and a random NP in the title respectively.

\paragraph{Adjacent Link}
The second baseline predicts the adjacent NP as a complement. We have two variants: predict the next NP as the complement (\textit{Adj-Forward}) or the previous NP (\textit{Adj-Backward}).

\paragraph{Surface Link}
The third baseline predicts surface links in the text, i.e., links in which the string ``\textbf{anchor} \underline{preposition} \textit{complement}'' appears as-is in the text. For instance, in ``Adam's father went to meet \textbf{the teacher} \underline{at} \textit{his school}'' it will predict the link (\textit{the teacher}, \underline{at}, \textit{his school}).
We also experiment with \textit{Surface-Expand}, a relaxed version which looks for the complement at a distance of up to 10 tokens following the anchor.

\paragraph{Combined} This baseline combines the three others, using the best strategy of each one (determined based on the empirical results), and predicts a link whenever at least one of the used baselines is triggered. Its purpose is to increase the recall.

\paragraph{Combined-Coref} This final baselines adds to the \textit{Combined} predictions the gold coreference information. For each link to an NP that is part of a coreference cluster, we also add links to all other NPs in the same cluster.

\subsection{Results}
The deterministic baselines' results are summarized in the first part of Table \ref{tab:results}.

In general, the F1 scores of the `single' baselines are low, ranging between 5.8 and 20.8 points, where the \textit{Adj-Backward} baseline achieves the lowest score and the \textit{Surface-Expand} baseline achieves the highest score. %
The \textit{Combined} baseline makes use of the best strategy of each previous baseline (based on the F1 score), that is, \textit{Title-Last}, \textit{Adj-Backward} and \textit{Surface-Expand}, and reaches 22.8 F1. \textit{Combined-Coref} extend the \textit{Combined} baseline by adding the coreference gold data, and achieves the best performance for the deterministic baselines, of an overall 25.2 F1.

These results demonstrate that (a) the links are spread across different locations in the text, and (b) that the data is unlikely to have clear shortcuts that models might exploit, while there are some strong structural cues.%

\input{tables/results}

\section{Neural Models}
\label{sec:neural}

\begin{figure}[t!]
\centering

\includegraphics[width=1.\columnwidth]{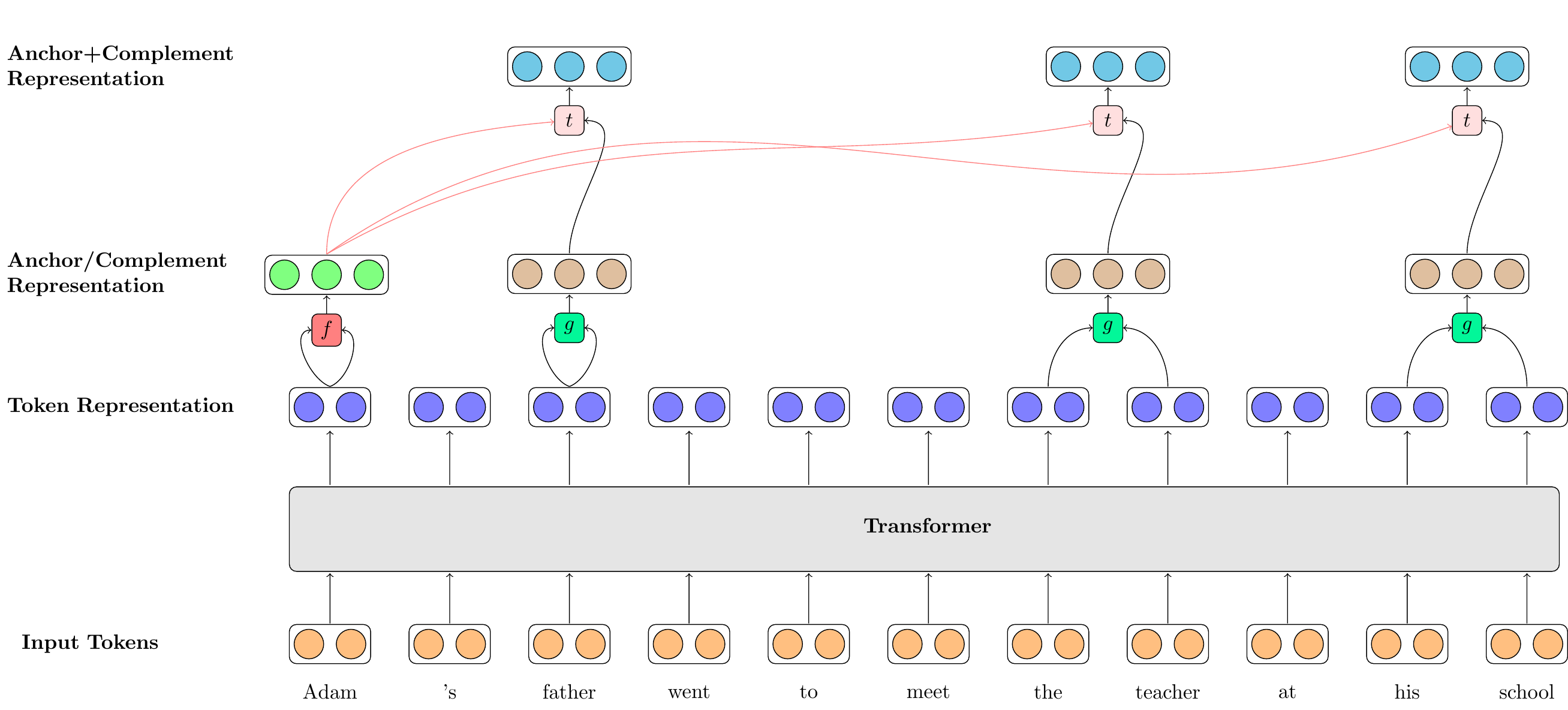}

\caption{A schematic view of the model's architecture.}
\label{fig:nn-architecture}
\vspace{-10mm}
\end{figure}

Next, we experiment with three neural models based on a pre-trained masked language model (MLM), specifically, SpanBERT \cite{spanbert}. We also experiment with an additional baseline with uncontextualized word embeddings.

\paragraph{Architecture} 
At a high level, our models take the encoding of two NPs --- an anchor and a complement --- and predict whether they are connected, and if so, by which preposition.
To encode an anchor-complement pair, we first encode
the text using the MLM and then encode each NP by concatenating the vectors of its first and last tokens. The resulting anchor and complement vectors are then each fed into a different MLP, each with a single 500-dimensions hidden-layer. The concatenation of the MLP outputs results in the anchor-complement representation. This representation is then fed into the prediction model, which has two variants. The architecture resembles the end-to-end architecture for modeling coreference resolution \cite{lee2017end}.
A schematic view of the architecture is presented in Figure \ref{fig:nn-architecture}.

\paragraph{Variants}
In the \textbf{decoupled} variant, we treat each prediction as a two-step process: one binary prediction head asks ``are these two NPs linked?'', and in case they are, another multiclass head determines the preposition.\footnote{During training, the preposition-selecting head is only used for NP pairs that are connected in the gold data.}
In the \textbf{coupled} variant, we have a single multiclass head that outputs the connecting preposition or \textsc{None}, in case the NPs are not connected.
We also experiment with a \textbf{frozen} (or ``probing'') variant of both models, in which we keep the MLM frozen, and update only the NP encoding and prediction heads. The frozen architecture is intended to quantify the degree to which the pretrained MLM encodes the relevant information, and it is very similar to the \emph{edge-probing} architecture of \citet{edge_probing}.
Finally, the \textbf{static} variant aims to measure how well a model can perform with NPs alone, without considering their context.
This model sums all the static embeddings of each span and uses the same modeling as the coupled prediction. This baseline uses the 300-dim word2vec non-contextualized embeddings \cite{w2v}. 
We experiment with two versions: decoupled and coupled.

\paragraph{Technical Details}
All neural models are trained using cross-entropy loss and optimized with Adam \cite{adam}, using the AllenNLP library \cite{allennlp}. 
We train using a $1e^-5$ learning rate for 40 epochs, with early stopping based the F1 metric on the development set.
We use SpanBERT \cite{spanbert} as the pretrained MLM, as it was found to work well on span-based tasks with its \emph{base} and the \emph{large} variants.
The anchor and complement encoding MLPs have one 500-dim hidden layer and output 500-dim representations. The prediction MLPs have one 100-dim hidden layer. All MLPs use the ReLU activation.
We used the same hyperparameters for all baselines and did not tune them.\footnote{Except for the static variant for which we also tried a larger learning rate of $1e^{-3}$, which worked better in practice.}

\subsection{Results}
\label{sec:results}

\paragraph{In-Domain Results}

The pretrained models are presented in the second part of Table \ref{tab:results}.
Overall, the fully-trained transformers in the \textit{coupled} variant perform significantly better than all other models, achieving 49.2 and 52.4 F1 in the base and large variants.
Interestingly, the static and frozen variants perform similarly: the F1 scores range between 15.1 and 23.2. It is worth noting that the static variant achieves better results than the frozen one. 
This corroborates our hypothesis that many of the capabilities needed to solve the task are not explicitly covered by the language-modeling objective and that the NPs information alone is not sufficient to solve the task, as was also argued in \citet{hou2020bridging,Probing-for-Bridging-Inference}.
Finally, we note an interesting trend that the decoupled variant favors recall whereas the coupled variant favors precision, across all models.
In summary, all models perform substantially below human agreement, leaving a large room for improvement.

\paragraph{OOD Results}

\input{tables/ood_results}

Here we report the best model's results (coupled-large) on the OOD data. The results are summarized in Table \ref{tab:ood_results}. We break down the results per domain (and per forum in the case of Reddit), as well as the human agreement results for comparison.
We observe a substantial drop in performance, with a large difference between domains (e.g. the model achieves on the IMDB split an overall 36.9 F1, while on Reddit - 28.2 F1). While the agreement scores for these domains are also lower than for the in-domain test set (88.6 F1),\footnote{The annotation of the Wikinews domain went on for a long time, which allowed for more training, revision-and-feedback loops and refinement of guidelines with focus on this specific type of texts and their challenges. This explains the somewhat higher agreement scores for this domain.}
the model's performance decreases more drastically on these splits.

\section{Analysis}
\label{sec:res-analysis}

\subsection{Quantitative Analysis}
\label{sec:quantitative-analysis}

\paragraph{Unlabeled Accuracy and Preposition-only Accuracy}
\input{tables/results_analysis}

To disentangle the ability to identify that a link exists between two NPs from the ability to assign the correct preposition to this link, we report also unlabeled scores (ignoring the preposition's identity) and preposition-only scores. 
IPrep-Acc is the accuracy of predicting the correct preposition over gold relations (NP pairs) where the unlabeled relation was correctly identified by the model. UPrep-Acc is the accuracy of predicting the correct preposition for gold NP pairs that were not identified by the model. 
The results (Table \ref{tab:results_analysis}) reveal a big gap between IPrep and UPrep accuracies for all models, indicating that the models are significantly better (yet far from perfect) at choosing the correct preposition when they identify that a relation should exist between two NPs. Overall, the preposition selection accuracy is significantly better than the majority baseline of choosing ``of'' for all cases (which would yield 23.5\%) but substantially worse than the human agreement which is almost 100\%. We also observe that while the unlabeled relation scores are indeed better than their labeled counterparts, the link-identification aspect of the task is significantly more challenging than choosing the correct preposition once the link was identified.

\paragraph{Preposition Analysis}

\begin{figure}[t!]
\centering

\includegraphics[width=1.\columnwidth]{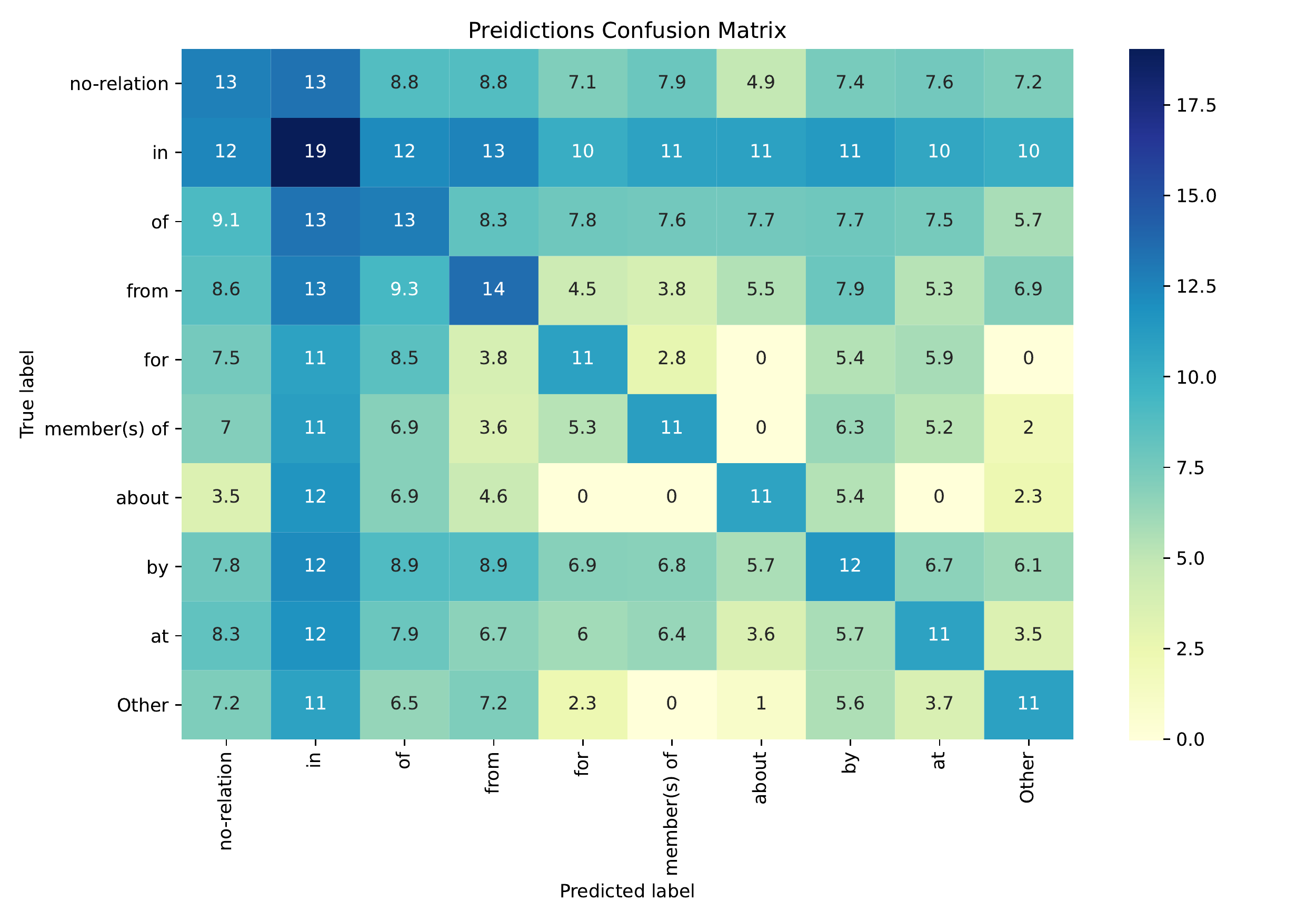}

\caption{A confusion matrix of the predictions of the Joint-large model over the test set. The numbers are in log2 scale (except for zero values which are untouched). We show show the 10 most common labels for brevity.}
\label{fig:confusion}
\vspace{-6mm}
\end{figure}

We analyze the errors of the best model on the different classes (the most common prepositions and  no-relation).
We present a confusion matrix in Figure \ref{fig:confusion}. The most confusing label is  \textit{in}, which is confused (both in false positive and false negative) with all  other labels. The preposition \textit{of}  is also confused quite frequently, while \textit{about} is confused much less.

\paragraph{Accuracy per NP Distance}

\begin{figure}[t!]
\centering

\includegraphics[width=.8\columnwidth]{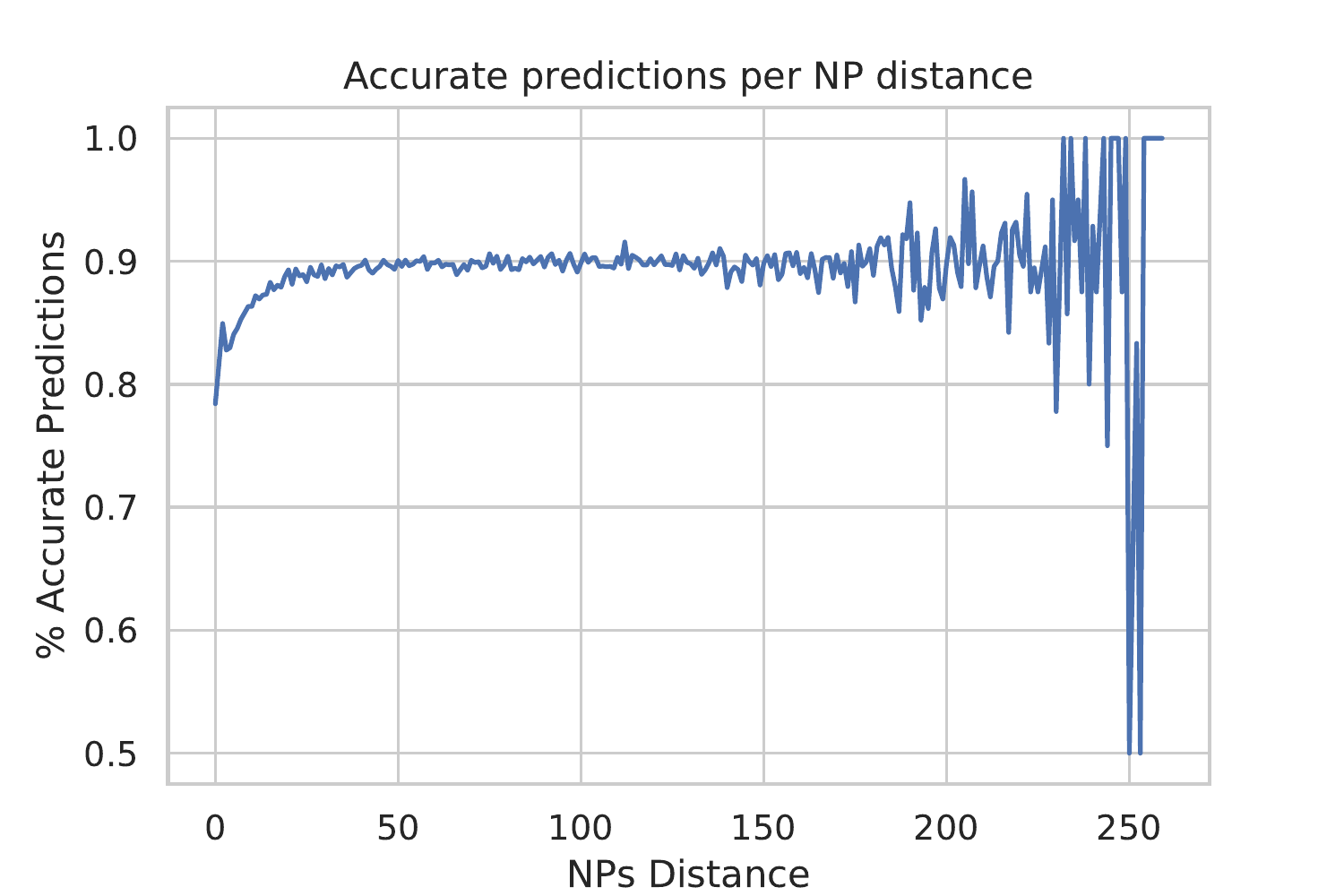}

\caption{Accuracy of the Joint-large model over the dev set, for every NP-distance bin.}
\label{fig:acc_per_dist}
\vspace{-6mm}
\end{figure}

We assess the effect of the linear distance between the two NPs on the ability of the model to accurately predict the link. For each NP pair in distance $x$, Figure \ref{fig:acc_per_dist} shows the percentage of correct predictions over that bin. We observe a trend of improved performance until 40 tokens, which then reaches a plateau of about 90\% (the results for distances above 180 are noisy due to data sparsity at these distances). Interestingly, the model struggles more in the short-distance links, rather than the ones farther apart.
We performed the same analysis on precision and recall errors, and found similar trends.

\subsection{Qualitative Analysis}

\begin{figure*}[t!]
\centering

\includegraphics[width=1.\textwidth]{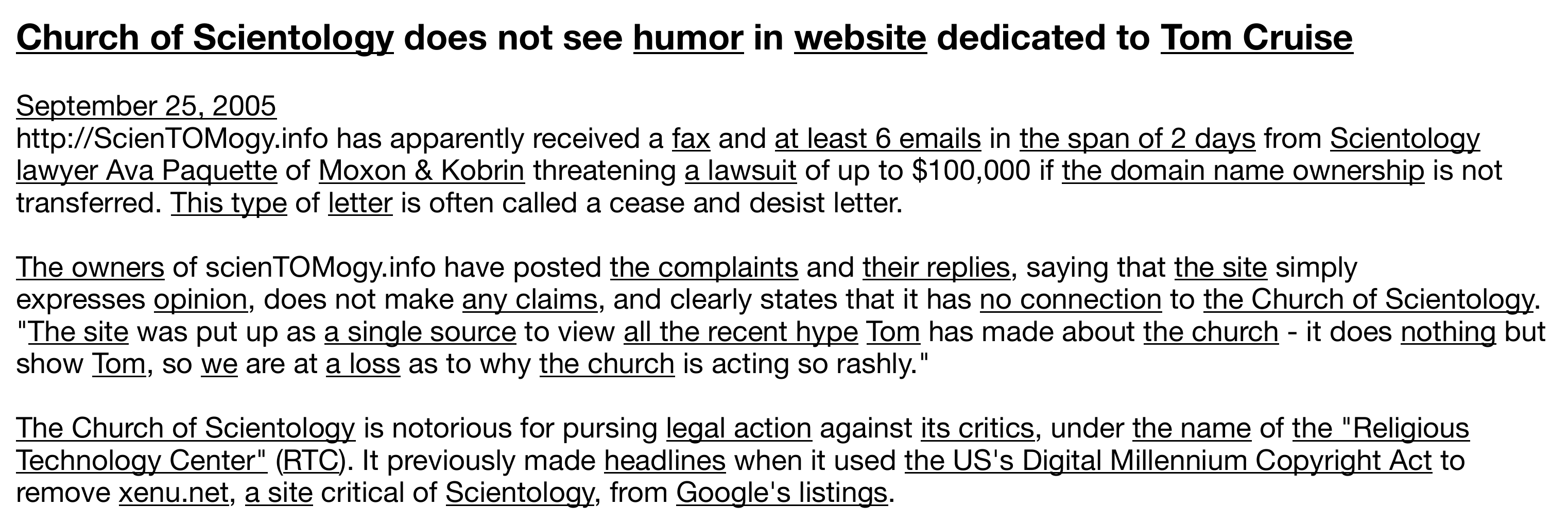}

\caption{Development-set document used for the qualitative error analysis. All 45 considered NPs are underlined.  Out of $45^2 - 45=1980$ potential links, this document contains 271 gold links, and 231 erroneously-predicted links, which we analyze.}
\label{fig:error-analysis-text}
\end{figure*}

\input{tables/error-analysis}

To better understand the type of errors, we zoom in on a single document (shown in Figure \ref{fig:error-analysis-text} and manually inspect all errors our best model (Coupled-large) made on it.

Out of the 1980 potential links, the model wrongly predicted 231 links (82 precision errors, where a model predicted an incorrect link, and 149 recall errors, where the model failed to identify a link). 
Out of the 231 disagreements with the gold labels, we found 84.2\% to indeed be incorrect, 10.5\% to actually be correct, and 5.3\% were found to be ambiguous.

Table \ref{tab:error-analysis} breaks down the errors into 9 categories, covering both type of errors and skills needed to solve them:
\textit{Preposition Semantics}: where the model predicted a link, but used a wrong preposition; \textit{Ambiguous}: where both gold and predicted answers can be correct, depending on the reading of the text; \textit{Wrong Label}: where the gold label is incorrect; \textit{Missing Label}: where the prediction was correct, as well as the original label, but the predicted preposition was missing (i.e. cases where more than one preposition are valid); \textit{Generics}: cases where the anchor is generic, and thus no link exists from it; \textit{Coreference Error}: where the model links to a complement that appears to be part of the coreference chain, but is not, or the annotator mistakenly attached an additional, erroneous NP to the chain; \textit{World Knowledge}: links that require some world knowledge in order to complete; \textit{Explicit}: where the link appears explicitly in the text, but the model did not predict it accordingly; and \textit{Other}, for none of the above.

In Table \ref{tab:results} we observed that the model is better at precision than recall. Here we also observe that recall and precision errors differ also in their type distribution.
In terms of precision, 17.0\% of the links were correct links with a wrong preposition. Such errors seem rather trivial, such that a good language model would not err: using an LM for explicitly quantifying the likelihood of links may be a promising direction for future work.
An interesting error that occurs both in the precision and recall errors is that of Ambiguous categorization --- for instance in the recall category, one interpretation can be read as an \textit{opinion} being expressed about \textit{Tom Cruise}, while the other interpretation reads \textit{opinion} in a more abstract way, thus not connected to Cruise.
Finally, the largest category in the precision errors and the most common category in recall errors is,  ``Other'', 
with varied mistake that do not single out noticable phenomena.

\section{Related  Tasks and Linguistic Phenomena}
\label{sec:related}

\input{tables/links_vs_bridging}

From the outset, recovering  NP-NP relations appears familiar from many previous linguistic endeavors. While 
TNE is related to them, it is certainly different, in scope, purpose and definition.

Our departure point for this work has been the notion of an {\em implicit argument of a noun}, i.e., nouns such as ``brother'' or ``price'' that are incomplete on their own, and require an argument to be complete. In linguistics, these are referred to as {\em relational nouns} \cite{partee1983a,relational-loebner, barker1995a,relational-de-bruin, Partee2000GenitivesRN,Lbner2015FunctionalCA,newell-cheung-2018-constructing}. %
In contrast, nouns like ``plant'', or ``sofa'' are called {\em sortal} and are conceived as ``complete''; their denotation need not rely on the relation to other nouns, and can be fully determined.

A sensible task, then, could be to identify all the relational nouns in the text  and recover their missing noun argument.
However, in practical terms, %
the distinction between sortal and relational %
is not clear-cut.
Specifically, sortal nouns  often stand in relations to other nouns, and these relations are useful for understanding the text and for fully determining the reference --- as in ``the sofa [\underline{in the house}]'', %
or  ``the sofa [\underline{on the carpet}]'' (as opposed to that   on the floor), and ``the house [\underline{of a particular owner}]''.

A closely related linguistic concept and an established  task 
 in the last decade  is {\em bridging anaphora resolution} \cite{bridging,loebner-relational-associative, poesio-vieira-1998-corpus,Matsui,Gardent,isnotes, hou-etal-2013-cascading,hou-etal-2013-global,nedoluzhko-2013-generic,hou-etal-2014-rule,grishina-2016-experiments,Rsiger2018BASHIAC,  hou-etal-2018-unrestricted,hou-2018-deterministic,hou-2018-enhanced, hou2020bridging,pagel-roesiger-2018-towards,roesiger-etal-2018-integrating,roesiger-2018-rule,Probing-for-Bridging-Inference,Kobayashi2021BridgingRM,Hou2021EndtoendNI}. Both bridging anaphora resolution and  \task{},  relate  entities mentioned in the text via non-identity relations. However, there are a number of major differences between bridging and  \task{}. 
These differences are summarized in Table \ref{tab:links-vs-bridging}, and expanded upon in what follows.

First, there is no agreed-upon definition of bridging  \cite{roesiger-etal-2018-bridging}.
Consequently,  manual annotation of bridging relations, and the use of these annotations,  requires substantial expertise and effort.
In contrast, \task{} is  compactly defined, and is amenable to large-scale annotation after only a brief annotator training.

Secondly, the relation between a bridging expression and its antecedent\footnote{In these studies the terms \textit{bridging expression} (or \textit{anaphoric NP}) and  \textit{antecedent} roughly correspond to our {\em anchor} and \textit{complement} respectively.}  has to be implicit. In \task{} the relations between the anchor and the complement are either implicit or explicit. 

Next, in most  bridging studies a bridge is a type of \textit{anaphora}: the bridging expression is not interpretable without the antecedent. In \task{} the anchor can in fact be interpretable on its own --- the complement supplements it with additional information (``sofa \underline{[on the carpet]}'') or simply exposes existing information in a uniform way.

Also, bridging expressions are not discourse-old, i.e., they can only refer to entities that are mentioned in the text for the first time. This implies that in a coreference chain only the first mention can have a bridging link. In \task{} there is no such restriction: an anchor can be either old or new. 
Furthermore, in many bridging works the antecedent does not have to be an NP. It can be also a verb or a clause. In \task{} both the anchor and the complement have to be NPs.

Finally, 
all the aforementioned studies have been defined by and written for linguists, using linguistic terminology, with a predominantly documentary motivation.
As a result the task definitions are often narrowly scoped,  highly technical and non-interpretable for non-experts --- making their annotation by crowd-workers essentially impossible. It also makes the consumption of the output by (non-linguist) NLP practitioners doubtful.

In this work we aimed to define a linguistically meaningful yet simple, properly scoped, and easy to communicate task. We want crowd-workers as well as downstream-task designers to be able to properly understand the task, its scope and its output, and we want the data collection procedure to be amenable to high inter-annotator agreement. %

\paragraph{A Note on  Decontextualization} 
 Recently,  \citet{decontextualization} 
 introduced the text-decontextualization task, in which the input is a text and an enclosing textual context, and the goal is to produce a standalone text that can be fully interpreted outside of the enclosing context. The decontextualization task involves handling multiple linguistic phenomena, and, in order to perform it well, one must essentially perform a version of the \task{} task. For example, decontextualizing ``Prices are expected to rise'' based on ``Temporary copper shortage. Prices are expected to rise'', involves establishing the relation ``Prices [\underline{of copper}] are expected to rise'').

Like our NP Enrichment proposal, the decontextualization task bears a strong application-motivated, user-facing perspective.  It is useful, well-defined and easy to explain. However, as it is entirely goal-based (``make this sentence standalone''), the scope of covered phenomena is somewhat eclectic. More importantly, the output of the decontextualization task is targeted at human readers rather than machine readers. For example, it does not handle relations between NPs that appear within the decontextualized text itself; it only recovers  relations of NPs with the surrounding context.
Thus, many implicit NP relations are left untreated.

\section{Conclusions}

We propose a new task named {\em Text-based NP Enrichment}, or TNE, in which we aim to annotate each NP with all its relations to other NPs in the text.
This task covers a lot of implicit relations that are nonetheless crucial for text understanding.
We introduce a large-scale dataset enriched with such NP links, containing 5.5K documents and over 1M links --- enough for training large neural networks --- and provide high-quality test sets, both in and out of domain.
We propose several baselines for this task and show that it is challenging --- even for state-of-the-art LM-based models --- and that there is a big gap from human performance.
We release the dataset, code, and models, and hope that the community will adopt this task as a standard component of the NLP pipeline. %
\newpage
\section*{Acknowledgements}
We would like to thank the NLP-BIU lab, Nathan Schneider, and Yufang Hou for helpful discussions and comments on this paper.
We also thank the anonymous reviewers and the action editors, Marie-Catherine de Marneffe and Mark Steedman, for their valuable suggestions.
Yanai Elazar is grateful to be supported by the PBC fellowship for outstanding PhD candidates in Data Science and the Google PhD fellowship.
This project has received funding from the European Research Council (ERC) under the European Union's Horizon 2020 research and innovation programme, grant agreement No.\ 802774 (iEXTRACT) and grant agreement No.\ 677352 (NLPRO).

\bibliography{tacl2018}
\bibliographystyle{acl_natbib}

\end{document}

%% file: tables/prepositions.tex
\begin{table}[t]
    \centering
\resizebox{1.\columnwidth}{!}{%
\begin{tabular}{l}
\toprule

\makecell[l]{of, against, in, by, on, about, with, after, to, \\from, for, among, under, at, between, during, \\near, over, before, inside, outside, into, around} \\

\bottomrule
\end{tabular}

}
\caption{Prepositions used in TNE.}
\label{tab:prepositions}

\vspace{-12mm}
\end{table}

%% file: tables/agreement.tex
\begin{table}[t]
    \centering
\resizebox{1.\columnwidth}{!}{%

\begin{tabular}{lrrrrrr|r}
            {} &       \multicolumn{2}{c}{In-Domain}       & \multicolumn{4}{c}{Out-of-Domain} & \\
            \cmidrule(lr){2-3} \cmidrule(lr){4-7}
               & train & test  & Books & IMDB & Reddit & OOD   & all \\
\midrule

CoNLL (Coref)    & -     & 82.1  & 76.8  & 77.6 & 78.6 & 77.1  & 79.8 \\

\midrule

Relation-F1  &   89.8 &  94.4 &   87.0 &  89.6 &    90.2 &     88.9 &  90.3 \\
IPrep-Acc &   99.8 & 100.0 &   99.5 &  99.8 &   100.0 &     99.8 &  99.9 \\
UPrep-Acc &  100.0 & 100.0 &  100.0 & 100.0 &   100.0 &    100.0 & 100.0 \\
\midrule
F1        &   89.6 &  94.4 &   86.6 &  89.4 &    90.2 &     88.6 &  90.1 \\

\bottomrule
\end{tabular}

}
\caption{Agreement scores on the different annotation parts. We report both the coreference CoNLL scores, and the metrics of \task{} calculated on the consolidated annotations.}
\label{tab:agreement}

\vspace{-10mm}
\end{table}

%% file: tables/stats.tex
\begin{table*}[t]
    \centering
\resizebox{1.\textwidth}{!}{%

\begin{tabular}{lrrrrrrr|r}
{} &     \multicolumn{3}{c}{In-Domain} & \multicolumn{4}{c}{Out-of-Domain} & \\
\cmidrule(lr){2-4} \cmidrule(lr){5-8}
{} &     train &       dev &      test &  Books &  IMDB &  Reddit &      OOD-all &         all \\
\midrule
Documents             &   3,988.0 &     500.0 &     500.0 &      170.0 &     169.0 &       170.0 &    509.0 &     5,497.0 \\
Tokens                & 651,835.0 &  81,741.0 &  77,618.0 &   30,133.0 &  29,285.0 &    27,181.0 & 86,599.0 &   897,793.0 \\
NPs                   & 143,406.0 &  17,815.0 &  17,521.0 &    6,502.0 &   6,099.0 &     5,803.0 & 18,404.0 &   197,146.0 \\
NP-Links              & 744,513.0 & 103,668.0 & 120,198.0 &   22,886.0 &  25,164.0 &    10,228.0 & 58,278.0 & 1,026,657.0 \\
Coref-Clusters        &  21,473.0 &   2,598.0 &   2,581.0 &      773.0 &     759.0 &       821.0 &  2,353.0 &    29,005.0 \\
Coref-Links           & 354,734.0 &  51,776.0 &  56,443.0 &   11,847.0 &  11,798.0 &     5,347.0 & 28,992.0 &   491,945.0 \\

\midrule

Avg. Surface Distance &      53.9 &      52.8 &      52.6 &       53.6 &      58.2 &        47.9 &     54.6 &        53.7 \\
Avg. Symmetric Links  &      10.2 &      12.6 &      14.3 &        6.7 &      26.5 &         1.7 &     11.6 &        10.9 \\
Avg. Transitive       &      95.5 &     119.2 &     144.2 &       52.8 &      64.5 &        14.8 &     44.0 &        97.3 \\

\midrule

\% Title Links         &      13.6 &      12.7 &      12.0 &        8.3 &      13.7 &        25.4 &     15.8 &        13.6 \\
\% Backward-relations  &      56.9 &      56.0 &      55.7 &       56.7 &      56.6 &        57.3 &     56.8 &        56.7 \\
\% Surface-Form        &       4.0 &       3.5 &       2.9 &        4.3 &       2.7 &         5.4 &      4.1 &         3.9 \\
\% Surface-Form+       &       6.3 &       5.6 &       4.6 &        6.8 &       3.5 &         7.2 &      5.9 &         6.0 \\
\% Intersentential  &       84.4 &       85.1 &       85.5 &        79.8 &       87.7 &         83.2 &      83.8 &         84.6 \\
\bottomrule
\end{tabular}

}
\caption{Statistics summary of the \dataset{} dataset.}
\label{tab:analysis}
\end{table*}

%% file: tables/results.tex
\begin{table}[t]
    \centering
\resizebox{1.\columnwidth}{!}{%

\begin{tabular}{clrrrrrr}
\toprule
& Model & Precision & Recall & F1 \\
\midrule

& Human*         &     94.8 &    94.0 & 94.4 \\

\midrule

 \parbox[t]{2mm}{\multirow{8}{*}{\rotatebox[origin=c]{90}{Deterministic}}} & 
Title-First    &       25.6 &        4.1 &         7.1 \\
& Title-Last     &       29.1 &        4.7 &         8.0 \\
& Title-Random   &       27.1 &        4.3 &         7.4 \\
& Adj-Forward   &       21.2 &        3.4 &         5.8 \\
& Adj-Backward  &       31.6 &        5.1 &         8.7 \\
& Surface        &       \textbf{43.5} &        3.3 &         6.2 \\
& Surface-Expand &       14.4 &       37.8 &        20.8 \\
& Combined       &       15.4 &       44.1 &        22.8 \\
& Combined-Coref &       16.4 &       \textbf{54.7} &        \textbf{25.2} \\

\midrule
\parbox[t]{2mm}{\multirow{10}{*}{\rotatebox[origin=c]{90}{Pretrained}}} & 
Decoupled-static & 10.1 & 58.8 & 17.2 \\
& Decoupled-frozen-base & 9.6 & 55.5 & 16.3 \\
& Decoupled-frozen-large & 9.7 & 56.2 & 16.5 \\
& Decoupled-base & 11.8 & 68.5 & 20.1 \\
& Decoupled-large & 12.0 & \textbf{69.9} & 20.5 \\

& Coupled-static & 59.6 & 14.4 & 23.2 \\ %
& Coupled-frozen-base & 60.1 & 8.6 & 15.1 \\
& Coupled-frozen-large & 58.4 & 11.5 & 19.2 \\
& Coupled-base & 60.4 & 41.5 & 49.2 \\
& Coupled-large & \textbf{65.8} & 43.5 & \textbf{52.4} \\
\bottomrule
\end{tabular}

}
\caption{Results of the deterministic baselines and neural models on the  test set. We report three metrics: the precision, recall and F1 of the overall relation predictions. The first row is an estimated human agreement on 10\% of the data, and not over the entire test set, thus marked with an asterisk.
Note that the first and second parts of the table are not directly comparable, since in the Deterministic results, the preposition labels is given by an oracle, whereas in the Pretrained results, it is predicted by the models.}
\label{tab:results}
\vspace{-3mm}
\end{table}

%% file: tables/ood_results.tex
\begin{table}[t]
    \centering
\resizebox{1.\columnwidth}{!}{%
\begin{tabular}{lrrr}
\toprule
Split &  Precision &  Recall &   F1 \\
\midrule
In-domain test  & 65.8 & 43.5 & 52.4 \\
\midrule
\midrule
Human*          &       80.5 &    93.7 & 86.6 \\
Books           &       46.3 &    28.4 & 35.2 \\

\midrule
Human*          &       89.8 &    89.0 & 89.4 \\
IMDB            &       51.7 &    28.6 & 36.9 \\

\midrule
Human*          &       91.2 &    89.3 & 90.2 \\
r/askedscience  &       48.3 &    25.7 & 33.6 \\
r/atheism       &       44.1 &    25.7 & 32.5 \\
r/LifeProTips   &       31.4 &    15.9 & 21.1 \\
r/AskHistorians &       36.4 &    26.0 & 30.3 \\
r/depressed     &       43.4 &    27.3 & 33.5 \\
r/YouShouldKnow &       36.5 &    20.9 & 26.6 \\
r/              &       37.8 &    22.5 & 28.2 \\

\midrule
Human*          &       86.9 &    90.5 & 88.6 \\
OOD             &       46.9 &    27.5 & 34.7 \\

\bottomrule
\end{tabular}

}
\caption{Results of the best model (Coupled-large) on OOD data, broken into the different sub-splits. The columns are the same as in Table \ref{tab:results}. Also reporting results on in-domain split for comparison.}
\label{tab:ood_results}

\vspace{-15mm}
\end{table}

%% file: tables/results_analysis.tex
\begin{table}[t]
    \centering
\resizebox{1.\columnwidth}{!}{%
\begin{tabular}{lrrrrr}
\toprule
{} & {Links-P} & {Links-R} & {Links-F1} & {IPrep-Acc} & {UPrep-Acc}  \\
\midrule

Human* &     94.8 &     94.0 &      94.4 & 100.0 &      100.0  \\

\midrule

Decoupled-static & 67.3 & 19.6 & 30.4 & 77.7 & 54.2 \\
Decoupled-frozen-base & 65.9 & 24.5 & 35.7 & 65.0 & 52.5 \\
Decoupled-frozen-large & 67.2 & 22.8 & 34.1 & 68.2 & 52.7 \\
Decoupled-base & 71.1 & 46.7 & 56.4 & 78.2 & 59.9 \\
Decoupled-large & 73.5 & 47.2 & 57.5 & 79.3 & \textbf{61.5} \\
Coupled-static  & 70.1 & 17.0 & 27.4 & \textbf{85.0} & 49.8 \\
Coupled-frozen-base & 73.8 & 10.6 & 18.5 & 81.5 & 44.0 \\
Coupled-frozen-large & 73.3 & 14.4 & 24.0 & 79.8 & 43.7 \\
Coupled-base & 76.4 & 52.4 & 62.2 & 79.1 & 50.7 \\
Coupled-large & \textbf{80.5} & \textbf{53.1} & \textbf{64.0} & 81.8 & 49.6 \\
\bottomrule
\end{tabular}

}
\caption{Additional metrics of the neural models on the TNE test set. We report five metrics: the precision, recall and F1 of the relation predictions, as well as the preposition accuracy on relations where the model predicted there's a relation (IPrep-Acc), as well as the accuracy where the model predicted there's no relation (UPrep-Acc). The first row is an estimated human agreement on 10\% of the data, thus marked with an asterisk. These results are comparable with the `Pretrained' part in Table \ref{tab:results}.}
\label{tab:results_analysis}

\vspace{-3mm}
\end{table}

%% file: tables/error-analysis.tex
\begin{table*}[t]
    \centering
\resizebox{1.\textwidth}{!}{%

\begin{tabular}{cllllll}
\toprule
& Error Type & \% (Number) & Anchor & Label & Complement & Prediction \\
\midrule

 \parbox[t]{2mm}{\multirow{7}{*}{\rotatebox[origin=c]{90}{Precision Errors}}} & 

Preposition Semantics & 17.0 (14) & \textbf{a lawsuit} & \underline{by} & \textit{Church of Scientology} & \underline{against} \\

& Ambiguous & 14.6 (12) & \textbf{opinion} & \underline{about} & \textit{Church of Scientology} & \underline{of} \\

& Wrong Label & 13.4 (11) & \textbf{a fax} & \underline{no-relation} & \textit{the Church of Scientology} & \underline{about} \\

& Missing Label & 13.4 (11) & \textbf{a fax} & \underline{to} & \textit{the site} & \underline{about} \\

& Generics & 12.1 (10) & \textbf{letter} & \underline{no-relation} & \textit{Church of Scientology} & \underline{to} \\

& Coreference Error & 7.2 (1) & \textbf{at least 6 emails} & \underline{no-relation} & \textit{Scientology} & \underline{about} \\

& Other & 21.9 (18) & \textbf{their replies} & \underline{no-relation} & \textit{Scientology} & \underline{to} \\

\midrule
\midrule

\parbox[t]{2mm}{\multirow{6}{*}{\rotatebox[origin=c]{90}{Recall Errors}}} &

World-Knowledge & 12.7 (18) & \textbf{Church of Scientology} & \underline{from} & \textit{US} & \underline{no-relation} \\

& Wrong Label & 8.5 (12) & \textbf{the complaints} & \underline{in} & \textit{a fax} & \underline{no-relation} \\

& Ambiguous & 4.9 (7) & \textbf{opinion} & \underline{about} & \textit{Tom Cruise} & \underline{no-relation} \\

& Explicit & 3.5 (5) & \textbf{Scientology lawyer Ava Paquette} & \underline{of} & \textit{Moxon \& Kobrin} & \underline{no-relation} \\

& Coreference Error & 2.8 (4) & \textbf{a fax} & \underline{to} & \textit{a single source} & \underline{no-relation} \\

& Other & 67.3 (95) & \textbf{Website} & \underline{about} & \textit{Tom} & \underline{no-relation} \\

\bottomrule
\end{tabular}

}
\caption{Error types, and their statistics, based on the text presented in Figure \ref{fig:error-analysis-text}. The first part of the table presents precision errors, where the model predicted some link considered to be an error. The second part presents recall errors, where the model predicted no link exists.}
\label{tab:error-analysis}
\end{table*}

%% file: tables/links_vs_bridging.tex
\begin{table*}[t!]

\centering
\resizebox{1.\textwidth}{!}{%
\begin{tabular}{p{5cm}|p{4cm}|p{4cm}|p{4cm}|p{4cm}}
\toprule
Description/Paper & \textbf{ISNotes} \cite{isnotes} & \textbf{BASHI} \cite{Rsiger2018BASHIAC} & \textbf{ARRAU} \cite{roesiger-2018-rule} & \textbf{TNE} (ours) \\ 
\midrule
The anchor/bridging expression can be \textbf{discourse-old}                                           & No & No & No & Yes \\ 
\midrule
The anchor/bridging expression has to be \textbf{anaphoric }(not interpretable without the antecedent) & Yes                                                                                                                                                                                                                                                                                                                                  & Yes                                                                                                                                                                                  & No. ``Most bridging links are purely lexical bridging pairs which are not context-dependent (e.g., Europe – Spain or Tokyo –Japan).''  \cite{hou2020bridging}                                                                         & No                                                                                                                                  \\ 
\midrule
\textbf{Cataphoric }links (to expressions that appear later in the text) are allowed                   & No                                                                                                                                                                                                                                                                                                                                   & No                                                                                                                                                                                   & No                                                                                                                                                                                                            & Yes                                                                                                                                 \\ 
\midrule
Links are annotated as \textbf{part of a larger task} (e.g. IS, anaphoric phenomena)                          & Yes                                                                                                                                                                                                                                                                                                                                  & No                                                                                                                                                                                   & Yes                                                                                                                                                                                                           & No                                                                                                                                  \\ 
\midrule
The relations in the links have to be \textbf{implicit}                                                & Yes                                                                                                                                                                                                                                                                                                                                  & Yes                                                                                                                                                                                  & Yes                                                                                                                                                                                                           & No                                                                                                                                  \\ 
\midrule
The relations in the links are \textbf{limited to certain sematic types}                               & No. Any relations are allowed, but, similarly \task{}, ``you must be able to rephrase the bridging entity by a complete phrase including the bridging entity and the antecedent.'' If the antecedent is an NP, rephrasals are restricted to a PP or possessive/Saxon genitive. ``Set bridging'' is allowed in special cases.             & No                                                                                                                                                                                   & Yes. Bridging is limited to a set of relations (part-of, element, subset, ``other'', ``undersp-rel'') \cite{arrau-article}. On the other hand, the ``undersp-rel'' category~ can include any relations. The relations are marked. & No. Any relations that can be expressed with a preposition, are included, as well as element-set and subset-set relations.  \\ 
\midrule
The antecedent/complement can be \textbf{not only nominal, but also verbal or clausa}l                 & Yes                                                                                                                                                                                                                                                                                                                                  & Yes                                                                                                                                                                                  & No. All bridging antecedents are nominal.                                                                                                                                                                                                       & No. Only nominal complements are included                                                                                           \\ 
\midrule
The bridging expression/anchor has to be \textbf{definite}                                             & No.                                                                                                                                                                                                                                                                                                                                  & No, but different labels are used to distinguish definite and indefinite expressions.                                                                                                & No                                                                                                                                                                                                           & No. Anchors can be both definite and indefinite                                                                                     \\ 
\midrule
\textbf{Multiple antecedents / complements} are allowed                                                      & Yes, but only if they have different mandatory roles in the argument structure of the bridging expression.                                                                                                                                                                                                                           & ``As a general principle, one antecedent has to be chosen. In special cases, e.g. comparative cases where two antecedents are needed, the annotator may create two or several links.''\footnote{See annotation guidelines for BASHI: https://www.ims.uni-stuttgart.de/documents/team/alt-nicht-mehr-da/roesigia/guidelines-bridging-en.pdf} & Multiple antecedents are not allowed by the guidelines but in practice  do occur in some cases where two antecedents appeared equally strong. However, such cases are being removed from ARRAU release 3 (forthcoming).                                                                                                                                                                                                     & All the complements of every anchor should be annotated. Multiple complements are allowed and very common.                          \\
\bottomrule
\end{tabular}
}
\caption{Bridging anaphora resolution vs. \task{} comparison.}
\label{tab:links-vs-bridging}
\end{table*}